\documentclass{article}
\usepackage{ijcai21}

\usepackage{times}
\usepackage{soul}
\usepackage{url}
\usepackage[hidelinks]{hyperref}
\usepackage[utf8]{inputenc}
\usepackage[small]{caption}
\usepackage{graphicx}
\usepackage{amsmath}
\usepackage{amsfonts}
\usepackage{amssymb}
\usepackage{mathtools} 
\usepackage{xcolor}
\usepackage{amsthm}
\usepackage{booktabs}
\usepackage{algorithm}
\usepackage{algorithmic}
\usepackage{subcaption}
\title{IJCAI--21 Formatting Instructions}

\title{Autoencoding Slow Representations for Semi-supervised Data Efficient Regression}

\author{
Oliver Struckmeier$^1$\footnote{Contact Author}\and
Kshitij Tiwari$^2$\And
Ville Kyrki$^1$\\
\affiliations
$^1$Intelligent Robotics Group, Aalto University, Finland\\
$^2$Perception Engineering Group, University of Oulu, Finland\\
\emails
oliver.struckmeier@aalto.fi,
kshitij.tiwari@oulu.fi,
ville.kyrki@aalto.fi
}

\begin{document}

\maketitle

\begin{abstract}
The slowness principle is a concept inspired by the visual cortex of the brain.
It postulates that the underlying generative factors of a quickly varying sensory signal change on a slower time scale.
Unsupervised learning of intermediate representations utilizing abundant unlabeled sensory data can be leveraged to perform data-efficient supervised downstream regression.
In this paper, we propose a general formulation of \textit{slowness for unsupervised representation learning} adding a slowness regularization term to the estimate lower bound of the $\beta$-VAE to encourage temporal similarity in observation and latent space.
Within this framework we compare existing slowness regularization terms such as the L1 and L2 loss used in existing end-to-end methods, the SlowVAE and propose a new term based on Brownian motion.
We empirically evaluate these slowness regularization terms with respect to their downstream task performance and data efficiency.
We find that slow representations lead to equal or better downstream task performance and data efficiency in different experiment domains when compared to representations without slowness regularization.
Finally, we discuss how the Fr\'echet Inception Distance (FID), traditionally used to determine the generative capabilities of GANs, can serve as a measure to predict the performance of pre-trained Autoencoder model in a supervised downstream task and accelerate hyperparameter search.
\end{abstract}

\section{Introduction}\label{sec:intro}
Learning representations that capture general high level information from abundant unlabeled sensory data remains a challenge for unsupervised representation learning.
Research in neuroscience suggests that a major difference between state-of-the-art deep learning architectures and the human brain is that cells in the brain do not react to single stimuli, but instead extract invariant features from sequences of fast changing sensory input signals~\cite{bengio2009slow}.
Evidence found in the hierarchical organization of simple and complex vision cells shows that time-invariance is the principle after which the cortex extracts the underlying generative factors of these sequences and that these factors usually change slower than the observed signal~\cite{wiskott2002slow,berkes2005slow,bengio2009slow}.

Computational neuroscientists have named this paradigm the \textit{slowness principle} wherein individual measurements of a signal may vary quickly, but the underlying generative features vary slowly.
For example, individual pixel values in a video change rapidly during short periods of time, but the scene itself changes slowly.
This principle has found application in \textit{Slow Feature Analysis} (SFA) as proposed in \cite{wiskott2002slow}.
SFA has shown promising results in computational neuroscience but little research has explored the possible applications of the underlying slowness principle to state of the art unsupervised representation learning methods.
Previous research has bridged this gap by employing temporal contrastive L1 and L2 losses in end-to-end tasks~\cite{mobahi2009deep,sermanet2018time,zou2012deep} such as classification and view-point invariant robot imitation learning.
Recently an unsupervised representation learning method, the SlowVAE~\cite{klindt2020towards}, has been leveraging the observed statistics of natural transitions in observation space to extended the VAE objective with a sparse temporal prior.

With this paper, we aim to provide a framework for learning slow representations in an unsupervised way using Variational Autoencoders.
The framework adds an interchangeable slowness loss term to the evidence lower bound (ELBO) of the VAE objective allowing us to investigate the effects of different slowness regularizers on the learned representations.
In addition to the L1 and L2 norm used in previous work and the slowness loss term proposed in \cite{klindt2020towards}, we propose a new slowness term based on a Brownian motion prior for latent space evolution which we call the S-VAE.
We empirically compare the $\beta$-VAE, the SlowVAE, S-VAE and $L1$/$L2$ slowness terms within our framework with respect to their performance and data efficiency on downstream few-shot regression tasks.
We find that although slow methods out-perform the $\beta$-VAE in both best case downstream performance and few-shot data efficiency, there is no universally superior slowness term.

This begs the question: how can one measure and predict general ``goodness'' of representations with respect to downstream tasks without having to evaluate performance on the downstream tasks directly.
To this end we conduct a large scale hyperparameter search and visualize latent space properties for a low dimensional problem.
Furthermore, we investigate quantitative measures for the quality of latent representations and find that the Fr\'echet Inception Distance as proposed by \cite{frechet} correlates with the downstream task performance\footnote{The FID measures the ability to generate realistic images from random samples of the latent distribution by comparing the distribution of the generated data to the true data distribution.}.
Being able to predict the downstream task performance, just by sampling from the latent distributions of the autoencoder, greatly accelerates the hyperparameter search during the unsupervised pre-training of the VAE and helps identifying good encoder model without the need to perform the downstream task.

The key \textbf{contributions} of this paper are as follows:
\begin{itemize}
    \item A general formulation of slowness in unsupervised representation learning
    \item A review of existing slowness models in context of the formulation
    \item Proposing a new slowness model S-VAE
    \item Empirical analysis of downstream task performance when learning from sparse labeled data using unsupervised pre-training with abundant unlabeled data
    \item Proposing the Fréchet-Inception Distance (FID) to be used as a metric to compare and predict downstream task performance of VAE models
\end{itemize}

\section{Preliminaries}
\subsection{Slowness Principle}
The slowness principle is based on the assumption that the true generative factors of a signal vary on slower time scales than raw sensory signals.
Research in computational neuroscience suggests that cell structures in the visual cortex have emerged based on the underlying principle of extracting slowly varying features from the environment\cite{berkes2005slow}.
Leveraging this principle, we can extract higher level invariant scene information which usually changes slower than for example the individual pixel values of a video.

The most well known application of the slowness principle is the slow feature analysis method (SFA) introduced in~\cite{wiskott2002slow}.
SFA is an unsupervised learning algorithm designed to extract linearly decorrelated features by expanding and transforming the input signal such that it can be optimized for finding the most slowly varying features from an input signal~\cite{wiskott2002slow}.
Extending the SFA method to nonlinear features has shown that the learned features share many characteristics with those of complex cells in the V1 cortex~\cite{berkes2005slow}.
Further applications of the slowness principle include transformation invariant object detection~\cite{franzius2011invariant}, pre-training of neural networks for improved performance on the MNIST dataset~ \cite{bengio2009slow} and the self organization of grid cells, structures in the rodent brain used for navigation~\cite{franzius2007grids,franzius2007slowness}.

\subsection{Contrastive Learning and the Slowness Principle}
The objective of contrastive learning is to encode observations and place them in a latent space using a metric score that allows to express (dis-)similarity of the observations.
Contrastive learning has been successfully applied in reinforcement learning~\cite{srinivas2020curl} and most recently for object classification in SimCLR~\cite{chen2020simple,chen2020big}.
These methods use a contrastive loss on augmented versions of the same observation, effectively learning transformation invariant features from images, and show that these representations benefit reinforcement learning and image classification tasks.

When using the time as the contrastive metric, similar to the slowness principle, we talk about \textit{time-contrastive} learning.
Time-contrastive learning has been applied successfully to learning view-point-invariant representations for learning from demonstration with a robot~\cite{sermanet2018time}.
Similar to our work,  in~\cite{mobahi2009deep}, used the coherence in video material to train a Convolutional Neural Network (CNN) for a variety of specific tasks.
While training two CNNs in parallel with shared parameters, in alternating fashion a labeled pair of images was used to perform a gradient update minimizing training loss followed by selecting two unlabeled images from a large video dataset to minimize a time-contrastive loss based on the $L1$ norm of the representations at each individual layer.
The experiments showed that supervised tasks can benefit from the additional pseudo-supervisory signal and that features invariant to pose, illumination or clutter can be learned.
Compared to ~\cite{mobahi2009deep} where a specific task is learned end-to-end with an additional supervisory signal, the slow methods presented in this paper decouple the unsupervised representation learning step from downstream task learning with the goal of learning task-agnostic representations.

Compared to the above methods, the GP-VAE proposed by \cite{fortuin2020gp} is a model for learning temporal dynamics for problems such as reconstructing missing input features especially in a medical context.
The core idea of the GP-VAE is to learn a latent embedding of high dimensional sequential data and to model latent dynamics using a Gaussian Process prior.
The authors claim this prior facilitates representations which are smoother and allow the reconstruction of missing features in the input space.
Compared to the GP-VAE, the methods presented in this paper address the problem of learning good representations for downstream tasks, instead of learning temporal dynamics of the signal.
Another noteworthy method for learning temporal dynamics is the temporal difference variational autoencoder (TD-VAE)~\cite{gregor2018temporal} which learns representations that encode an uncertain belief state from which multiple possible future scenarios can be rolled out.

\subsection{Disentanglement}
Another concept related to unsupervised representation learning, especially when talking about the $\beta$-VAE, is disentanglement.
In this paper disentanglement is not compared as a metric, but used when discussing the properties of latent representations.

Although there is no definition of disentanglement, most works agree \cite{bengio2013representation,locatello2019challenging,klindt2020towards} on the common notion that disentangled representations should approximate the ground truth generative factors of the observation data and the dimensions should be largely independent from each other.
Ideally each disentangled factor represents one ground truth factor that led to the generation of the observation data.
Disentanglement in the context of the $\beta$-VAE has been discussed more in depth by \cite{burgess2018understanding}.
In the $\beta$-VAE pressure on the latent bottleneck of the autoencoder limits how much information can be transmitted per sample while at the same time trying to maximize the data log likelihood.
This is done by enforcing a unit Gaussian prior on the latent distributions which results in the embedding of data points on a set of representational axes where nearby points on the axes are also close in data space.
This regularization results in these axes being the main contributors to improvements in the data log likelihood and therefore often coincide with the ground truth generative factors.
Some of the claimed benefits of disentangled representations are better downstream task data-efficiency and interpretability~\cite{scholkopf2012causal,bengio2013representation,peters2017elements}.
However, a in their research Locatello et Al. \cite{locatello2019challenging,locatello2019fairness} show that various disentanglement methods are not able to generate disentangled representations without implicit biases on model and data, and that more disentanglement does not necessarily lead to better downstream task data-efficiency.
In a later work \cite{locatello2020weakly} the aforementioned challenges were addressed and the authors showed that with weak supervision it is possible to learn fair and generalizable representations.

\subsection{Fr\'echet Inception Distance}
In this paper we will make use of the Fr\'echet Inception Distance (FID) which is briefly explained in this section.
The FID as introduced in \cite{frechet} is a measure for the generative capabilities of deep generative models commonly used to measure how similar the images generated by GANs are to the real data distribution.
The FID is an improvement of the the Inception Score (IS) introduced by \cite{inception} which only evaluates the distribution of the generated images and does not compare it to the true data distribution which has been shown to fail when comparing models \cite{inceptioncompare}.
It is computed by comparing the activation distributions of a Inception-v3 neural network pre-trained on the ImageNet dataset for the generated and true data using the Wasserstein-2 between the distributions $(\mu, C)$, a Gaussians with mean $\mu$ and covariance $C$ describing the generated data distribution, and $(\mu_w, C_w)$ describing the real data distribution.

\section{Slow Variational Autoencoding Framework}\label{sec:generaldef}
The variational autoencoder (VAE) introduced by \cite{kingma2013auto} is a representation learning method for dimensionality reduction using a loss on the reconstruction quality of observations decoded from low dimensional representations.

Let $q_\theta$ be the variational approximate posterior distribution obtained by an encoder network with parameters $\theta$ and $\mathbf{z}$ be the latent vector such that $\mathbf{z} \sim q_\theta(\mathbf{z|o})$ where $\mathbf{o}$ is the observation to be reconstructed.
The decoder network denoted by $p_\phi(\mathbf{o|z})$ is parameterized by $\phi$.
Since it is computationally not tractable to directly maximize the log probability of the data, the estimate lower bound (ELBO) $\mathcal{L}$ is used for optimization:
\begin{equation}
    \max_{\theta,\phi} \log p_\phi (\mathbf{o}) \geq \mathcal{L} = \mathcal{L}_\text{rec} - \mathcal{L}_\text{ib}
    \label{eq:bvae_highlevel}
\end{equation}
where $\mathcal{L}_\text{rec} = \mathbb{E}_{q_\theta(\mathbf{z|o})}[\log p_\phi (\mathbf{o}|\mathbf{z})]$ is the reconstruction loss between the true observation and the decoded observations which is to be maximized.
$\mathcal{L}_\text{ib} =  D_{KL}(q_\theta(\mathbf{z|o})||p(\mathbf{z}))$ is constraining the information bottleneck by imposing a unit Gaussian prior and is promoting disentanglement in the latent space.
To make the sampling process differentiable (and thus trainable using gradient based optimization), the variational distribution is usually reparametrized as a Gaussian $q_\theta = \mathcal{N}(\mu, \Sigma)$.

\cite{higgins2016beta} introduced the $\beta$-VAE by adding a parameter $\beta$ to constrain the weight of the KL divergence between the prior and the posterior to allow a trade-off between disentanglement of the latent factors and reconstruction quality.

We extend the formulation of \cite{higgins2016beta} by an additional slowness constraint which is added to the constrained optimization problem as
\begin{equation}
\begin{split}
\mathcal{L}_\text{rec} \text{  subject to  } &\mathcal{L}_\text{ib} < \epsilon_1, \\
& \mathcal{L}_\text{slow}  < \epsilon_2
\end{split}
\label{eq:constrained}
\end{equation}
Rewriting Eq. \ref{eq:constrained} under the KKT conditions results in
\begin{equation}
    \mathcal{F} = \mathcal{L}_\text{rec} - \beta(\mathcal{L}_\text{ib}-\epsilon_1) - \lambda(\mathcal{L}_\text{slow} - \epsilon_2)
    \label{eq:slow_intermediate}
\end{equation}

With the simplification of $\beta, \epsilon_1, \lambda, \epsilon_2 \geq 0$, we can rewrite the above equation as free energy objective function
\begin{equation}
    \mathcal{F} \geq \mathcal{L} = \mathcal{L}_\text{rec} - \beta\mathcal{L}_\text{ib} - \lambda\mathcal{L}_\text{slow}.
    \label{eq:slow_highlevel}
\end{equation}
with $\beta$ being the $\beta$-VAE parameter and $\lambda$ being a parameter to control the weight of the slowness regularization.

Compared to disentanglement in the $\beta$-VAE, slow latent methods assume temporal continuity in both training data and downstream task as the underlying rule to represent the data in latent space.
The slowness principle here serves as a bias applied to the model and training data that is grounded in neuroscientific findings as opposed to the concept of disentanglement which is based on the assumption that independent factors are generally good, which has been subject to criticism lately~\cite{locatello2019challenging}.

\section{Slowness loss terms}
The slowness loss term $\mathcal{L}_\text{slow}$ is an  interchangeable component used to apply the slowness principle to the $\beta$-VAE formulation.
Let $D=(\tilde{o}_1, \ldots, \tilde{o}_T)$ be a sequence of unlabeled data points $\tilde{o}=(\mathbf{o},t)$ consisting of an observation $(\mathbf{o})$ and the time index $(t)$.
The index is used to determine the metric $\Delta t$ that for how far apart a pair of observations at times $i,j$ are, where $i<j$.
Specifically,
\begin{equation}
\Delta t(\tilde{\mathbf{o}}_i, \tilde{\mathbf{o}}_j) = j - i.
\end{equation}

\subsection{Lp-norm Slowness}
The most obvious way to describe $\mathcal{L}_\text{slow}$ is to take a $L_p$-norm of the mean $\mathbf{\mu}_i$ and $\mathbf{\mu}_j$ of two encoded latent distributions from two different points in time
\begin{equation}
    \mathcal{L}_\text{slow}(\mathbf{\tilde{o}}_i, \mathbf{\tilde{o}}_j) = (\mathbf{\mu_i-\mu_j})^p
\end{equation}
where $q_\theta(\mathbf{z|o}) = N(\mu, \Sigma)$.

This way of enforcing temporal similarity has been explored in \cite{mobahi2009deep} where a $L1$ norm was used to enforce similarity between two halfs of a siamese CNN architecture to leverage temporal similarity of observations during training.
Other use cases of a $L_p$-norm can be found in \cite{zou2012deep} and \cite{sermanet2018time}, where  a $L1$ norm and a triplet loss (based on L2 norm) respectively were used to achieve viewpoint invariance by enforcing similarity of views form different angles according to temporal similarity.
\cite{cadieu2012learning} claim that there are no significant differences between $L1$ and $L2$ norm for enforcing slowness in latent space.

While the above works show the benefits of $L1$ and $L2$ norm based slowness models, they are learning the task end-to-end and do not take the uncertainty $\Sigma$ of the latent distributions into account.

\subsection{SlowVAE}
The SlowVAE by \cite{klindt2020towards}, is based on a study of the statistics of natural transitions of object and mask positions in two acknowledged video-object segmentation datasets which follow generalized Laplace distributions.
The SlowVAE uses these Laplace distributions with a kurtosis $\alpha < 2$, assuming that the temporal transitions are sparse.
By formulating the SlowVAE ELBO in the proposed framework we get
\begin{equation}
\label{eq:SlowVAELoss}
\begin{split}
\mathcal{L}(&\mathbf{\theta, \phi, \beta, \lambda, o_i, o_{i+1}}) = \\
& \mathbb{E}_{q_\theta(\mathbf{z_i,z_{i-1}|o_i, o_{i-1}})}[\log p_\phi(\mathbf{o_i, o_{i-1}|z_i, z_{i-1}})]\\
& - \beta D_{KL}(q_\theta(\mathbf{z_{i-1}|o_{i-1}})||p(\mathbf{z_{i-1}}))\\
& - \gamma \mathbb{E}_{q_\theta(z_{i-1}|o_{i-1})}[D_{KL}(q_\theta(\mathbf{z}_i|\mathbf{o}_i))|p(\mathbf{z}_i|\mathbf{z}_{i-1})).
\end{split}
\end{equation}
where $\beta=1$ and corresponds to the KL loss term in the variational autoencoder.
The parameter $\gamma$ is a weight for the slowness loss term which is the KL divergence between approximate posterior at the current time and a factorized generalized Laplace distribution describing the prior on the sparse temporal transitions $p(\mathbf{z}_i|\mathbf{z}_{i-1})$\footnote{Following the general notation from Eq. \ref{eq:slow_highlevel} we will refer to this weight as $\lambda$ from here on.}.
The expectation is taken over the posterior of the previous step.

Compared to SVAE, by taking the expectation over the slowness KL-term, the SlowVAE compares a Gaussian centered at $\mu_i$ to a Laplace distribution at $\mu_{i-1}$ while in SVAE both terms are Gaussian. Moreover, in SVAE the slowness KL-term considers both covariances $({\Sigma}_i, {\Sigma}_{i-1})$ while in SlowVAE only ${\Sigma}_i$ is considered.

\subsection{S-VAE}
We now describe our proposed Slow Variational Autoencoder (S-VAE).
Considering two distinct yet sequential observations $\mathbf{o}_i,\mathbf{o}_j \in D~|~i<j$, the difference of the corresponding latent representations is given by the approximate difference distribution,
\begin{equation}
q_\theta(\mathbf{z}_j-\mathbf{z}_i|\mathbf{o}_j,\mathbf{o}_i) = \mathcal{N}(\mathbf{\mu}_j-\mathbf{\mu}_i, \Sigma_j + \Sigma_i) \equiv q_\theta(\Delta \mathbf{z}|\mathbf{o}_j,\mathbf{o}_i).
\end{equation}

To express the decaying similarity with growing temporal separation $\Delta t$, we assume a prior that the latent vector exhibits Brownian motion.
This allows the S-VAE to benefit from the closed form solution of the Brownian motion and to handle varying sampling intervals as opposed to the SlowVAE which does not incorporate the length of time interval in the formulation.
Moreover, all additive stochastic processes with stationary uncertainty approach Brownian motion in the limit.
Denoting by $Z_i, Z_j$ two increments of the Brownian motion at times $i$ and $j$, respectively, the \textbf{prior} distribution  $p(\Delta\mathbf{z})$ is given by
\begin{equation}
Z_{j} - Z_{i} = \sqrt{\Delta t} \cdot N \sim \mathcal{N}(0, \Delta t I) \equiv p(\Delta\mathbf{z}),
\end{equation}
where $N \sim \mathcal{N}(0, I)$.
This prior also encodes the time, resulting in observations further apart in time to have a prior distribution with a larger covariance.

The similarity of two observations can be computed as the Kullback-Leibler (KL) divergence between the approximate posterior and the prior distributions as
\begin{equation}
\label{eq:similarityloss}
\mathcal{L}_\text{slow}(\mathbf{\tilde{o}}_i, \mathbf{\tilde{o}}_j) = D_{KL}(q_\theta(\Delta\mathbf{z}|\mathbf{o}_j,\mathbf{o}_i)||p(\Delta\mathbf{z})).
\end{equation}

$\Delta t$ can be considered a scaling factor for the variance of the Brownian motion.
Thus, when considering two consequent elements in the sequence, we want to constrain $\mathcal{L}_\text{slow}$ to be smaller than a certain bound. 
We consider only pairs of consecutive elements because the bound (scaled according to $\Delta t)$ becomes weaker as the temporal distance between elements increases.
Using the slowness loss formulation form Eq. \ref{eq:slow_highlevel} we arrive at the S-VAE optimization target
\begin{equation}
\label{eq:SVAEloss}
\begin{split}
\mathcal{L}(&\mathbf{\theta, \phi, \beta, \lambda, o_i, o_{i-1}}) = \\
& \mathbb{E}_{q_\theta(\mathbf{z_i,z_{i-1}|o_i, o_{i-1}})}[\log p_\phi(\mathbf{o_i, o_{i-1}|z_i, z_{i-1}})]\\
& - \beta D_{KL}(q_\theta(\mathbf{z_i|o_i})||p(\mathbf{z}))\\
& - \lambda D_{KL}(q_\theta(\Delta\mathbf{z}|\mathbf{o}_{i},\mathbf{o}_{i-1})||p(\Delta\mathbf{z})).
\end{split}
\end{equation}
where $\theta$ and $\phi$ parameterize encoder and decoder and the parameters $\beta$ and $\lambda$ are hyperparameters for the strength of the disentanglement and slowness regularization.
The reconstruction loss term describes how well $\mathbf{o}_i$ can be reconstructed given its latent representation $\mathbf{z}_i$.
The disentanglement term is the Kullback-Leibler Divergence between the latent representation $\mathbf{z}_i$ and unit Gaussian prior $p(\mathbf{z})$.

\section{Empirical comparison}
In this section, we compare the previously introduced slow autoencoding methods to the $\beta$-VAE with respect to their performance when learning downstream regression tasks.
Although the downstream task performance benefits of disentangled representations are still subject of discussion, we include the $\beta$-VAE as a baseline in our comparison as it is one of the most established unsupervised representation learning methods.
Lastly, we will investigate the influence of the latent bottleneck term of the $\beta$-VAE and the slowness enforced by $\lambda$ on the representations and the downstream task performance.

Although the use case is different, we compared the TD-VAE to the slow methods and the $\beta$-VAE and found that it is outperformed by all methods.
We did not include those results as we were not confident in their thoroughness.
The problem being that to our knowledge there is no official code repository for the TD-VAE and we were not able to reproduce the results on the more complex DeepMind Lab experiment with the given implementation instructions.

\subsection{Experiment Setting}
Three experiments have been conducted in which the goal was to learn downstream tasks in a semi-supervised way from abundant unlabeled video data and sparse labeled data.
The semi-supervised process consists of a two step approach, first an autoencoding model is trained to extract representation from the abundant unlabeled data in an unsupervised way.
Second, a downstream task is learned using the encoded latent representations from the pre-trained models while keeping the encoder network weights frozen.
The downstream tasks involve temporal tasks like estimating odometry or velocity of an object from consecutive frames.
The performance is evaluated by observing the downstream task performance for varying amounts of labeled data.

The three experiments, described in more detail in Appendix \ref{app:experiments}, were conducted in different domains, a simple synthetic dataset of a ball bouncing in a $2D$ arena, two agents playing the game of Pong and a reinforcement learning agent exploring a $3D$ world in the DeepMind Lab environment~\cite{beattie2016deepmind}.
\begin{figure}[H]
\vskip 0.2in
\begin{center}
\centerline{\includegraphics[width=\columnwidth]{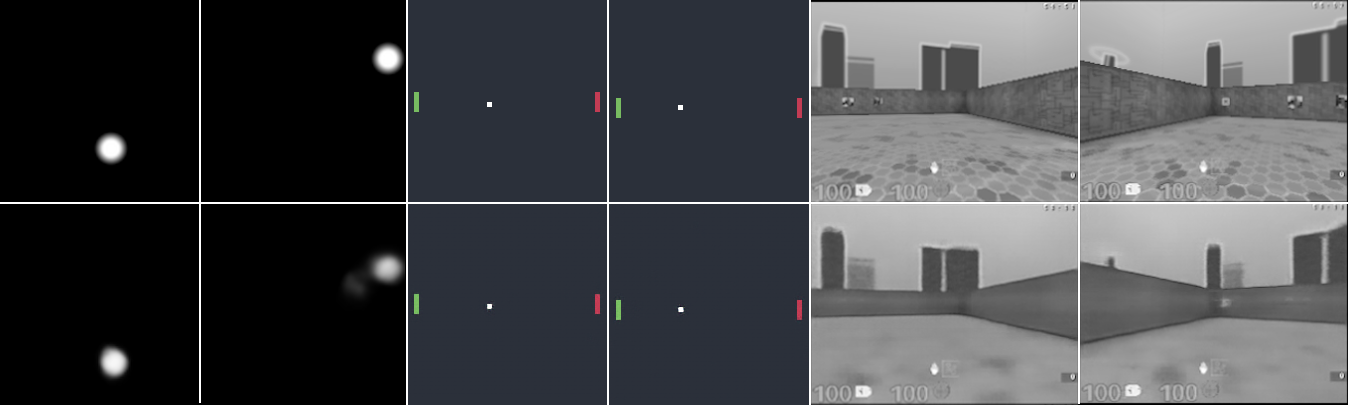}}
\caption{Images (top) and their reconstructions (bottom) obtained from the S-VAE. Two random samples of the full dataset for each of the three datasets.}
\label{fig:observations}
\end{center}
\vskip -0.2in
\end{figure}
Figure \ref{fig:observations} shows random observations of all three environments and reconstructions generated using the S-VAE.
\begin{figure*}
\begin{subfigure}{0.33\linewidth}
\begin{center}
\includegraphics[width=\textwidth]{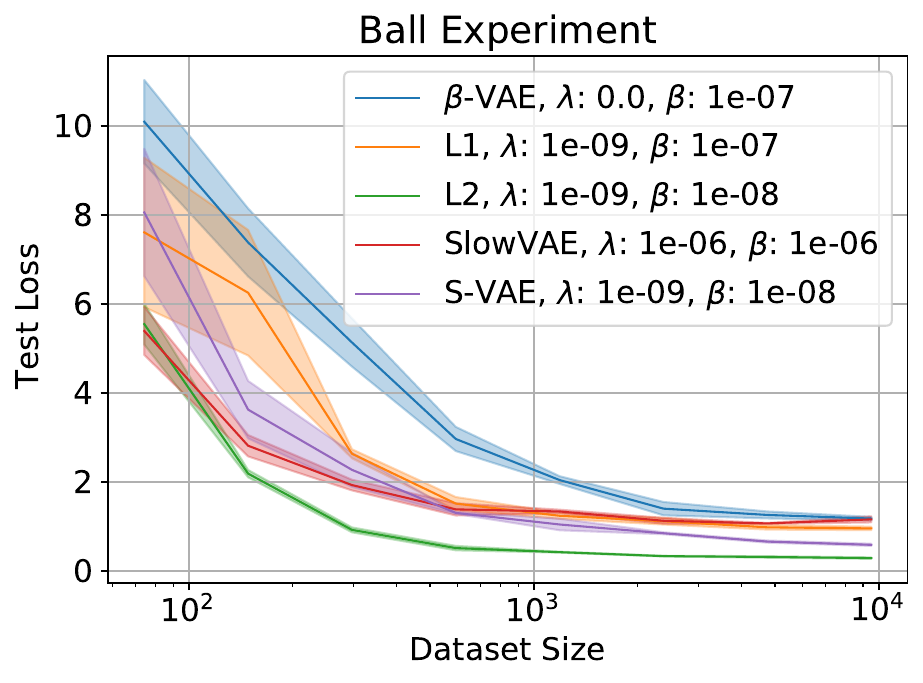}
\end{center}
\caption{Mean Squared Error Loss between true and predicted velocity vector vs. unseen test dataset size for the Ball experiment.}
\label{fig:performance_ball}
\end{subfigure}
\begin{subfigure}{0.33\linewidth}
\begin{center}
\includegraphics[width=\textwidth]{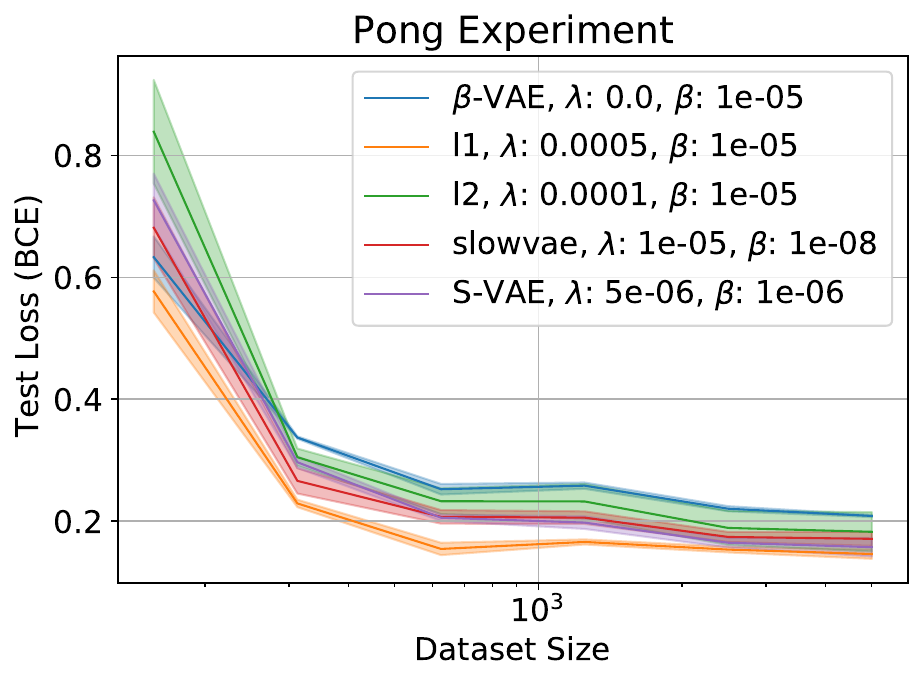}
\end{center}
\caption{Binary Cross Entropy Loss between true and predicted action probabilities vs. unseen test dataset size for the Pong experiment.}
\label{fig:performance_pong}
\end{subfigure}
\begin{subfigure}{0.33\linewidth}
\begin{center}
\includegraphics[width=\textwidth]{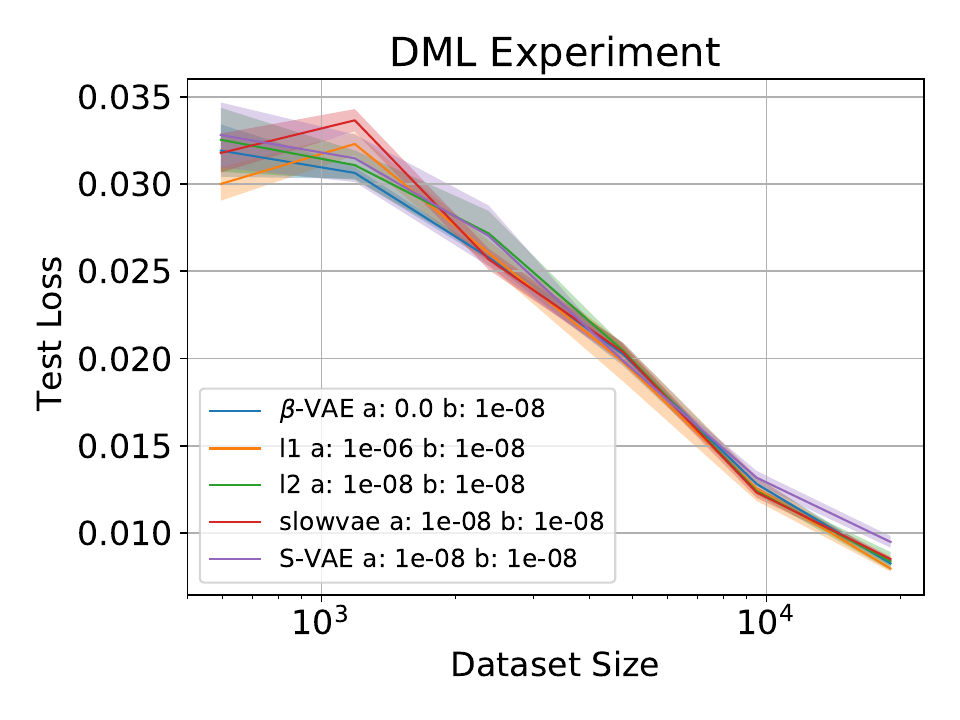}
\end{center}
\caption{MSE Loss between true and predicted movement vector vs. unseen test dataset size for the DeepMind Lab experiment.}
\label{fig:performance_dml}
\end{subfigure}
\caption{Average loss of all three experiments with standard deviation on unseen test dataset over 3 runs with different random seeds compared to the labeled dataset size used during the few-shot learning.}
\label{fig:performances}
\end{figure*}
\subsection{Evaluation of Downstream Task Performance}
The downstream task performance is measured by computing the loss on a previously unseen labeled test set.
During this analysis we will look at two main aspects of the downstream performance.
First, the \textit{best case performance} which refers to the downstream task performance given abundant labeled data.
Second, the \textit{data efficiency} measured by the downstream task performance for sparse labeled data.
\begin{figure*}[t]
\begin{subfigure}{0.24\linewidth}
\begin{center}
\includegraphics[width=\textwidth]{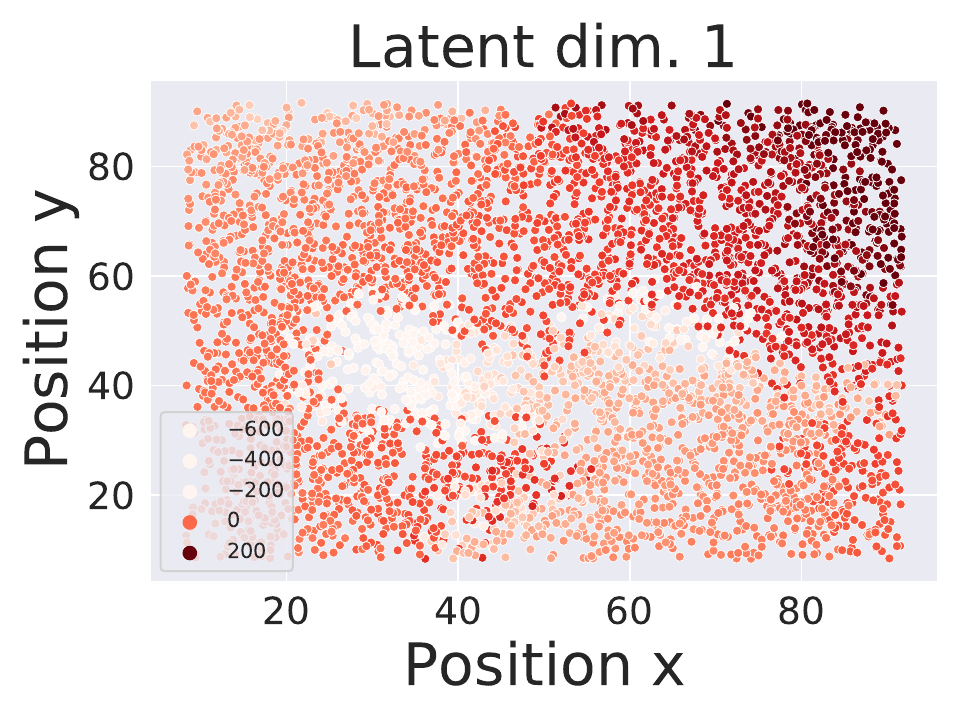}
\end{center}
\caption{$\beta$-VAE ($\lambda=0.0$).}
\label{fig:2d_latent_bvae}
\end{subfigure}
\begin{subfigure}{0.24\linewidth}
\begin{center}
\includegraphics[width=\textwidth]{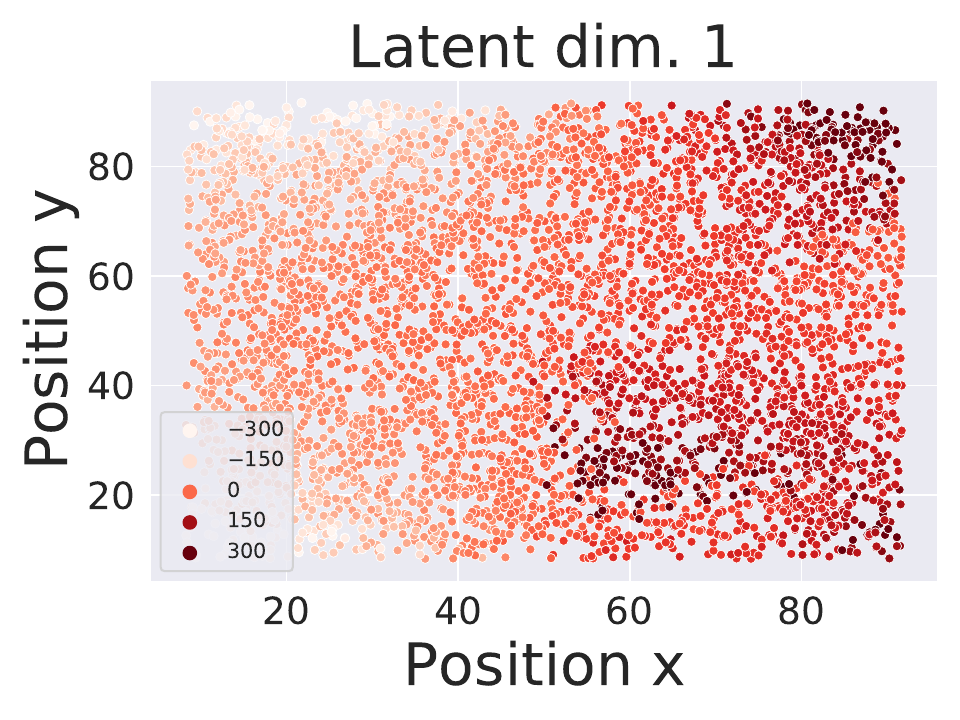}
\end{center}
\caption{S-VAE $\lambda=1e-09$.}
\label{fig:2d_latent_svae_1}
\end{subfigure}
\begin{subfigure}{0.24\linewidth}
\begin{center}
\includegraphics[width=\textwidth]{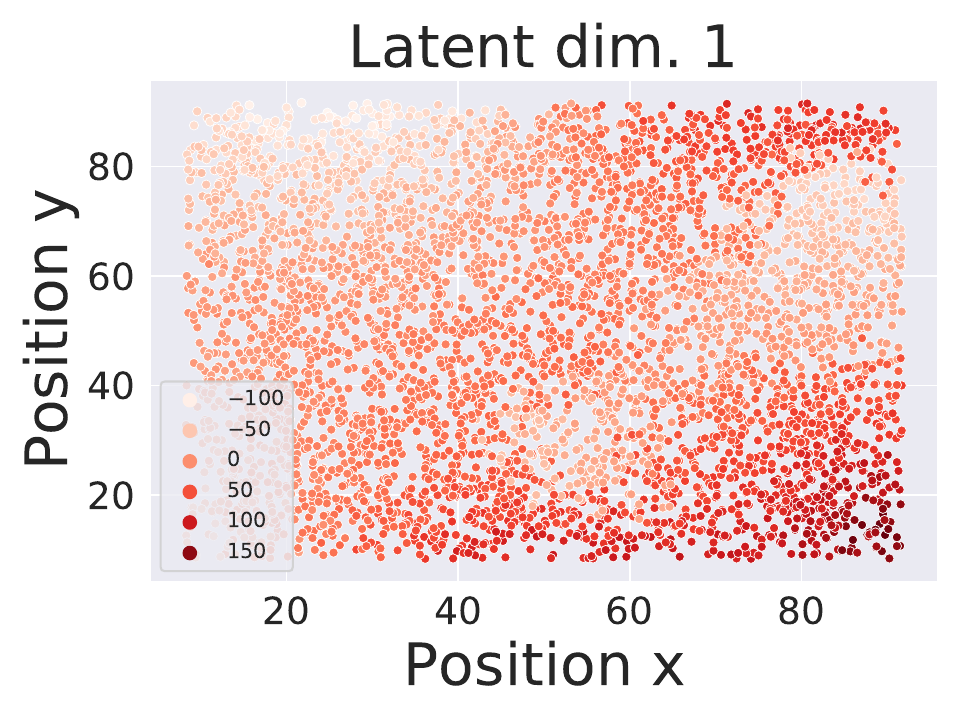}
\end{center}
\caption{S-VAE $\lambda=1e-06$.}
\label{fig:2d_latent_svae_2}
\end{subfigure}
\begin{subfigure}{0.24\linewidth}
\begin{center}
\includegraphics[width=\textwidth]{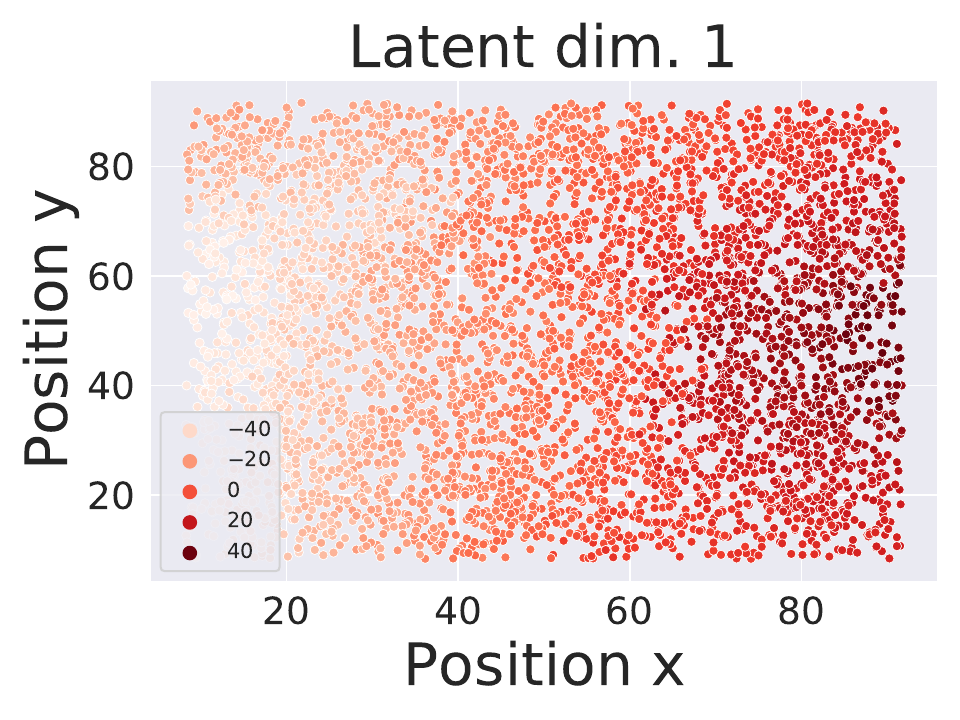}
\end{center}
\caption{S-VAE $\lambda=1e-04$.}
\label{fig:2d_latent_svae_3}
\end{subfigure}
\caption{Scatter plot visualization of 1500 samples in the $2D$ latent space of the Ball experiment. X- and Y-Axis are the position of the ball in the $100\times100$ pixel large environment and the color scale represents the value of the corresponding two latent dimensions. Fixing $\beta$ at $1e-08$ and varying $\lambda$ shows that as the latent space becomes visibly smoother.}
\label{fig:2d_latents}
\end{figure*}
Fig. \ref{fig:performances} shows the \textbf{best performing model} for each method as test loss given different amounts of labeled data.
We can see that in the ball and pong experiments the slow methods consistently outperform their $\beta$-VAE counterparts in both best case performance and data efficiency.
The $L2$ slowness loss method in the ball experiment achieved $75\%$ better best case performance (lower downstream task loss) and achieved similar performance as the $\beta$-VAE with only $2.73\%$ of the data showing significant benefits.
In the ball experiment, the best performing method used the $L2$ loss term and achieved $30\%$ better best case performance and performed as good as the $\beta$-VAE method with only $7.5\%$ of the data.
In the DeepMind Lab experiment all methods performed similarly across all dataset sizes. This is likely due to the hyperparameter only partially exploring the optimal hyperparameter region.
Some evidence for this hypothesis can be seen from Fig. \ref{fig:dml_overview_full} in Appendix \ref{app:analysis} where the best performing methods are clustered in the bottom right corner hinting that smaller values for $\lambda$ and $\beta$ would likely have yielded better performance.
The \textbf{mean performance} of all models trained with a given slowness loss term shows $L1$ and $L2$ loss performed similar compared to each other and better than the SlowVAE or the SVAEin many cases in all three experiments.
This indicates that in our experiments the slow methods do not necessarily benefit from taking the covariances of the latent distributions into account.
The finding about the $L1$ and $L2$ methods is in line with the findings in \cite{cadieu2012learning} claiming that there is no significant difference between $L1$ and $L2$ temporal loss terms.
More details and the average performance numbers can be found in Appendix \ref{app:performance}.

\subsection{Influence of Hyperparameters}
\label{sec:hyperparams}
Next we investigate how the hyperparameters $\lambda$ and $\beta$ influence the latent representations by visualizing the $2D$ latent space used in the ball experiment.
Fig. \ref{fig:2d_latents} shows a scatter plot where the x- and y-axis are the ground truth position of the ball in the arena and the color represents the latent value.
For simplification we only show latent dimension $1$ for each autoencoder model.
From left to right the slowness regularization is increased from $0$ (equivalent to $\beta$-VAE) to $1e-04$ (strongest regularization).
We can see that from left to right the continuity of the latent space increases.
For the beta-VAE abrupt changes of latent values create steep valleys in the latent space whereas for the S-VAE model with increasing slowness regularization the latent space becomes almost perfectly smooth from left to right ($\lambda = 1e-04$).
The S-VAE method with $\lambda = 1e-09$ and $\beta = 1e-08$ shown in Fig. \ref{fig:2d_latent_svae_1} is the overall best performing S-VAE method shown in Fig. \ref{fig:performance_ball}.
This indicates that the strongest slowness regularization does not necessarily lead to better downstream task performance, however, having both slowness and disentanglement regularization performs better than only disentanglement.
Looking at Figures \ref{fig:ball_overview_full}, \ref{fig:ball_overview_part}, \ref{fig:pong_overview_full}, \ref{fig:pong_overview_part} and \ref{fig:dml_overview_full}, \ref{fig:dml_overview_part} in the Appendix \ref{app:analysis}, this effect is highlighted for the S-VAE for all three experiments, strongest regularization (highest $\lambda$ or $\beta$) does not lead to best performance.

\begin{figure*}[t]
\centering
\begin{subfigure}{0.24\linewidth}
\begin{center}
\includegraphics[width=\textwidth]{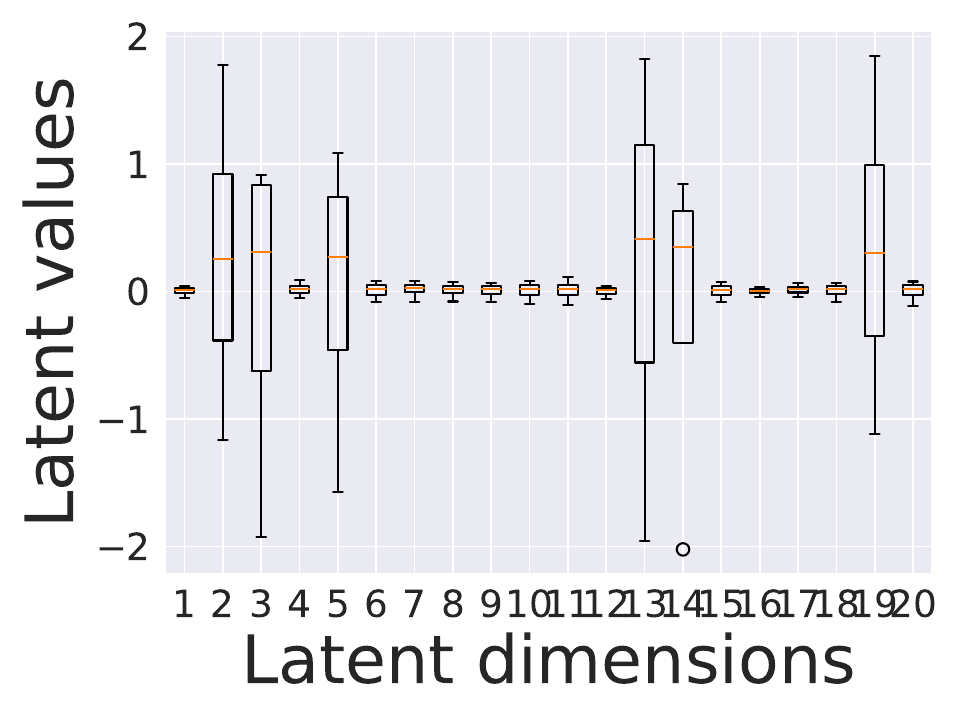}
\end{center}
\caption{Pong experiment, latent distributions with $\lambda = 1e-04$ and  $\beta$-VAE with $\beta = 1e-04$.}
\label{fig:pong_overview_1}
\end{subfigure}
\begin{subfigure}{0.24\linewidth}
\begin{center}
\includegraphics[width=\textwidth]{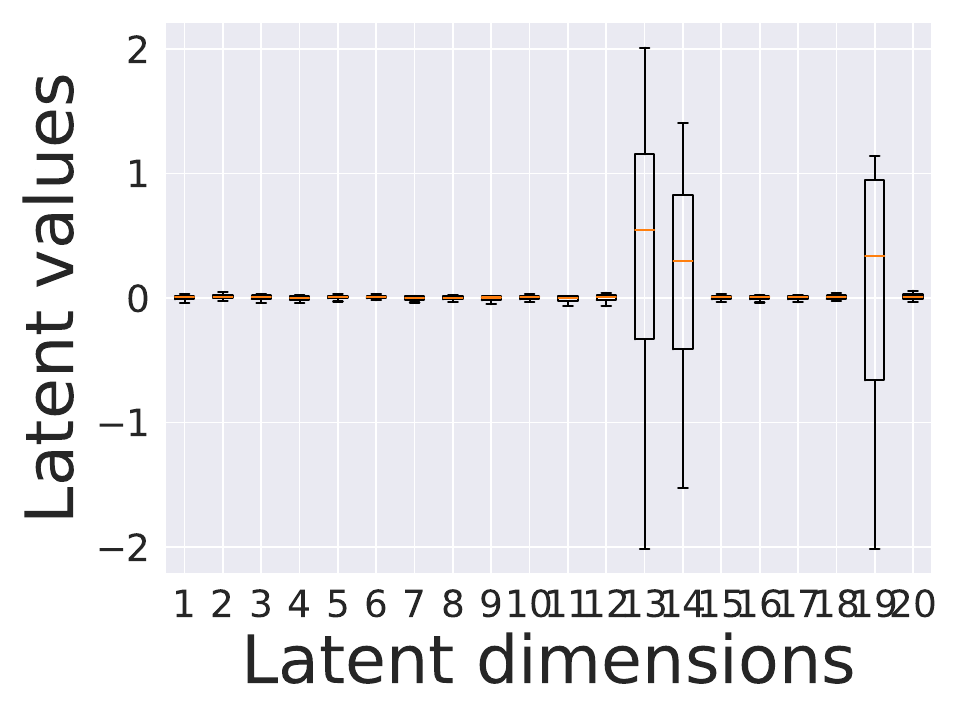}
\end{center}
\caption{Pong experiment, latent distributions with $\lambda = 1e-07$ and  $\beta$-VAE with $\beta = 1e-04$.}
\label{fig:pong_overview_2}
\end{subfigure}
\begin{subfigure}{0.24\linewidth}
\begin{center}
\includegraphics[width=\textwidth]{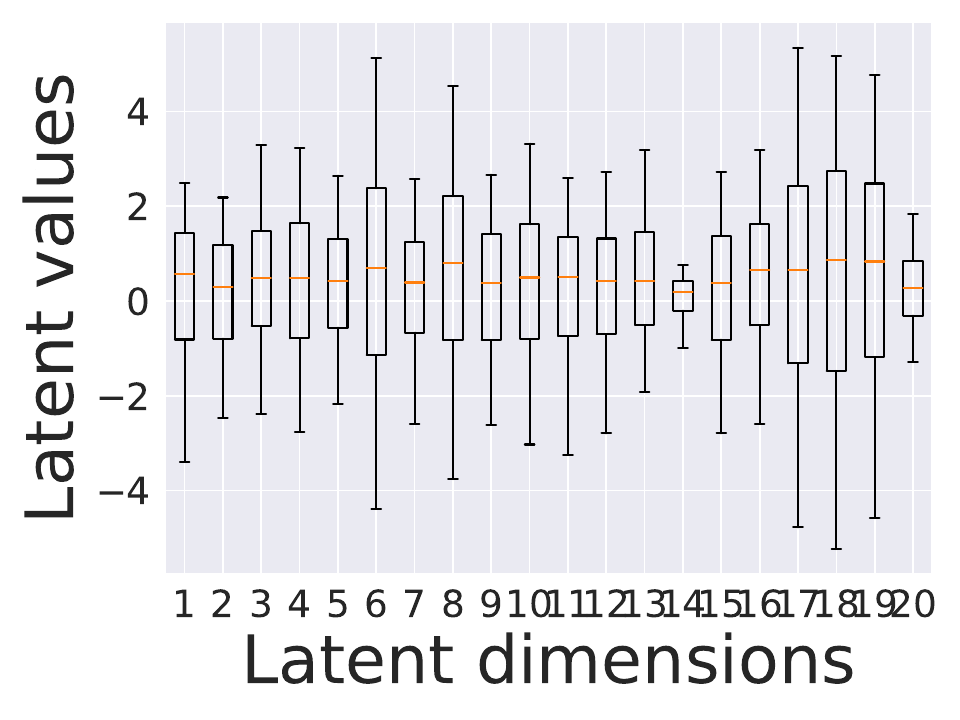}
\end{center}
\caption{Pong experiment, latent distributions with $\lambda = 1e-04$ and $\beta$-VAE with $\beta = 1e-06$.}
\label{fig:pong_overview_3}
\end{subfigure}
\begin{subfigure}{0.24\linewidth}
\begin{center}
\includegraphics[width=\textwidth]{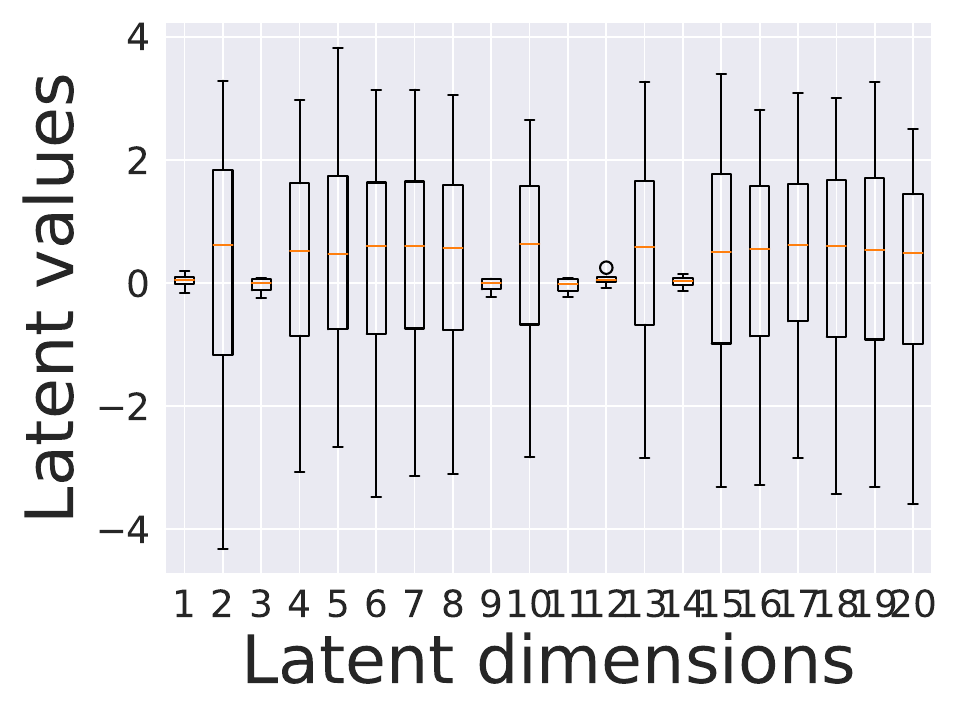}
\end{center}
\caption{Pong experiment, latent distributions with $\lambda = 1e-07$ and  $\beta$-VAE with $\beta = 1e-06$.}
\label{fig:pong_overview_4}
\end{subfigure}
\caption{Latent space distributions of two S-VAE models in the Pong experiment. Increasing $\lambda$ or $\beta$ reduces the number of latent channels used.}
\label{fig:pong_overview}
\end{figure*}

To demonstrate the influence of the hyperparameters $\lambda$ and $\beta$ on the latent channels, we visualize the $20$ latent dimensions of the pong experiment for each model in Fig. \ref{fig:pong_overview} and in more detail for the S-VAE in Fig. \ref{fig:latentanalysis}.
Latent dimensions encoding no information have mean and variance close to zero.
Accordingly, a threshold on the variance for each latent distribution is selected to compute the number of used dimensions.
As expected, higher values for $\beta$ and therefore stronger regularization reduce the number of used latent dimensions.
On the same note it can be hypothesized that stronger slowness regularization leads to more pressure on the latent channels and therefore reduce the number of used dimensions.
However, the data highlighted in Fig. \ref{fig:pong_overview_1} and \ref{fig:pong_overview_3} and in more detail in Fig. \ref{fig:pong_overview_used} and \ref{fig:dml_overview_used} shows multiple configurations of higher $\lambda$ where the number of used dimensions increases, thus providing evidence that the hypothesis does not hold.
Further investigation is needed to explore this phenomenon.
Looking at Figures
\ref{fig:ball_overview_full}, \ref{fig:ball_overview_part}, \ref{fig:pong_overview_full} and \ref{fig:pong_overview_part}, we observe no correlation between performance and used dimensions.

Although the qualitative continuity of the latent space and the number of dimensions used help us understand the effects of the hyperparameters $\beta$ and $\lambda$ on the latent space, it is not sufficient to reliably predict downstream task performance.
\begin{figure*}
\begin{subfigure}{0.32\linewidth}
\begin{center}
\includegraphics[width=\textwidth]{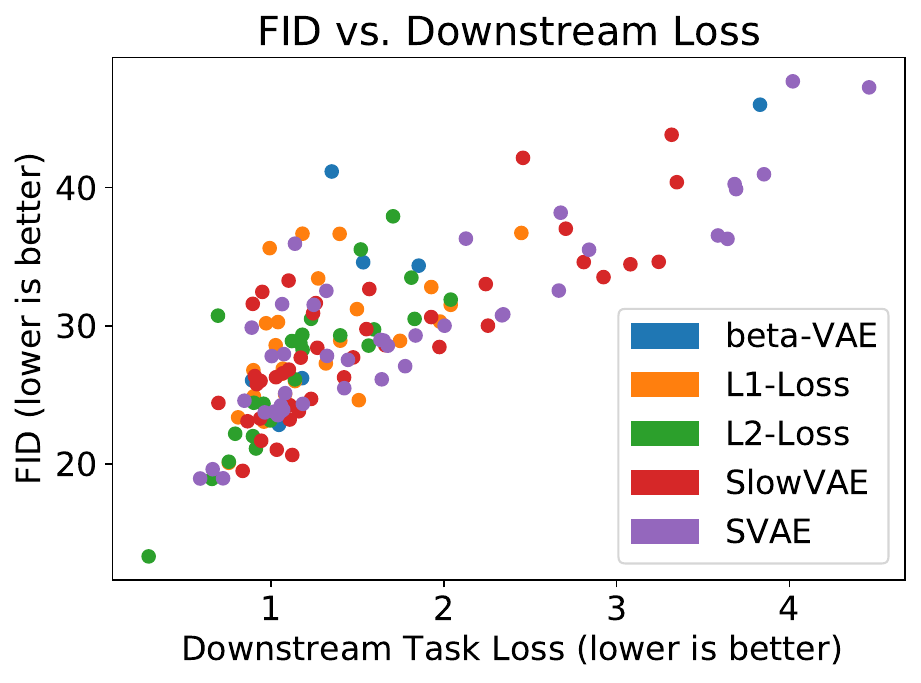}
\end{center}
\caption{Ball, full dataset}
\label{fig:blob_fid_full}
\end{subfigure}
\begin{subfigure}{0.32\linewidth}
\begin{center}
\includegraphics[width=\textwidth]{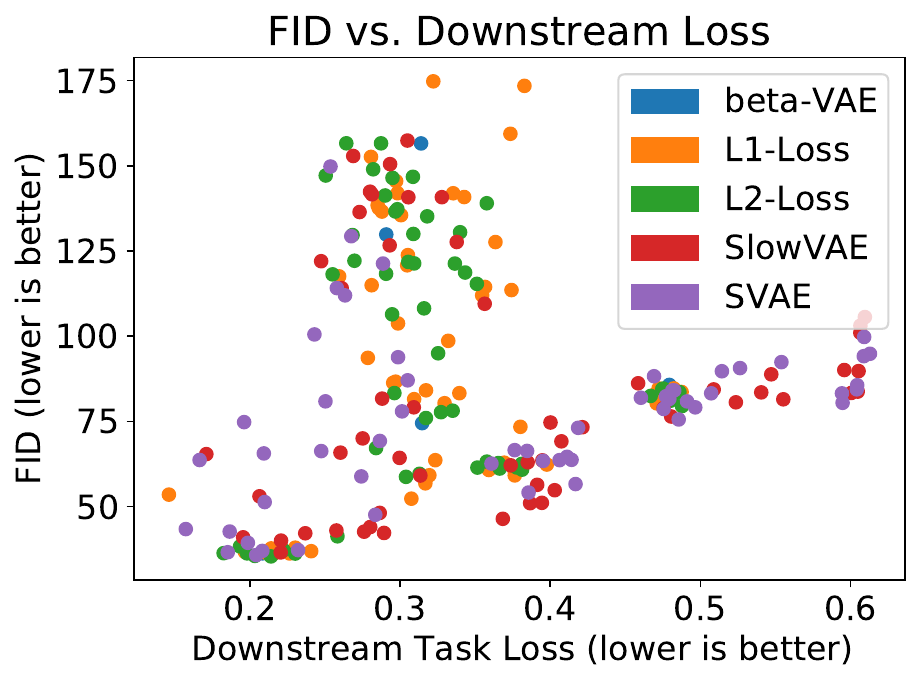}
\end{center}
\caption{Pong, full dataset}
\label{fig:pong_fid_full}
\end{subfigure}
\begin{subfigure}{0.32\linewidth}
\begin{center}
\includegraphics[width=\textwidth]{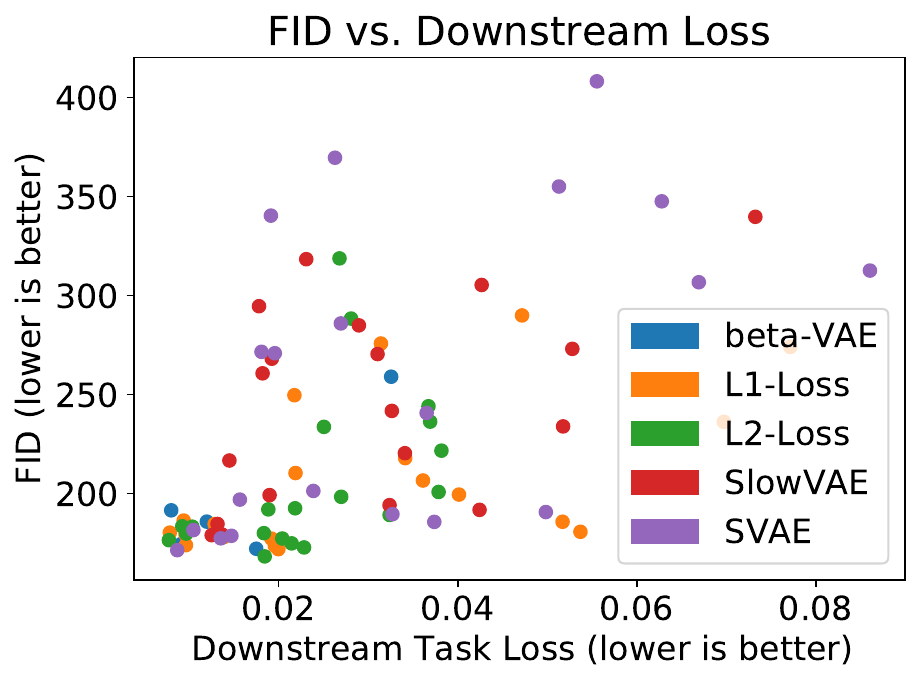}
\end{center}
\caption{DeepMind Lab full dataset}
\label{fig:dml_fid_full}
\end{subfigure}
\\
\begin{subfigure}{0.32\linewidth}
\begin{center}
\includegraphics[width=\textwidth]{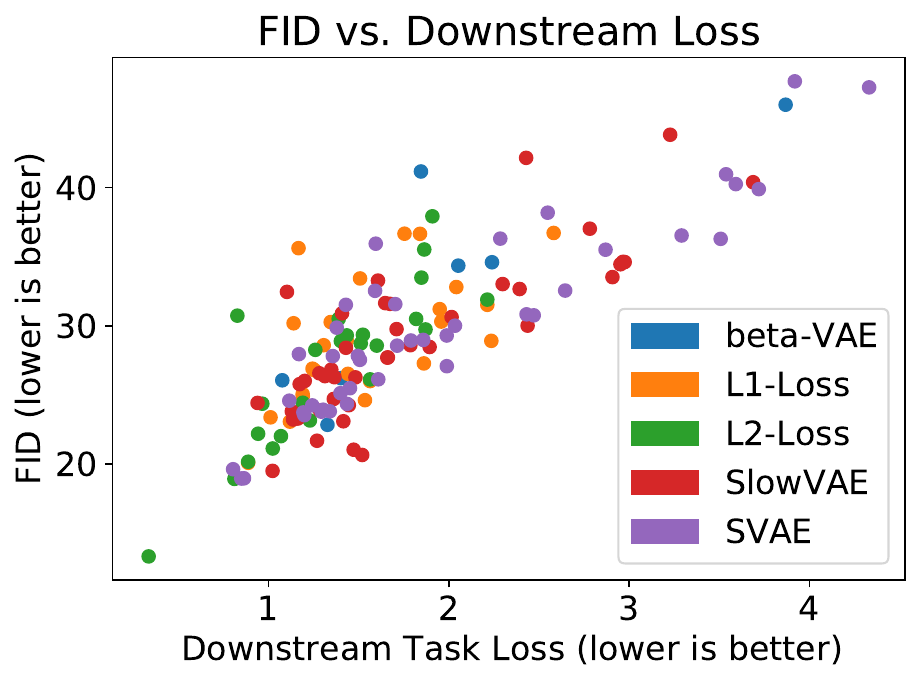}
\end{center}
\caption{Ball, $\frac{1}{4}$ of data}
\label{fig:blob_fid_part}
\end{subfigure}
\begin{subfigure}{0.32\linewidth}
\begin{center}
\includegraphics[width=\textwidth]{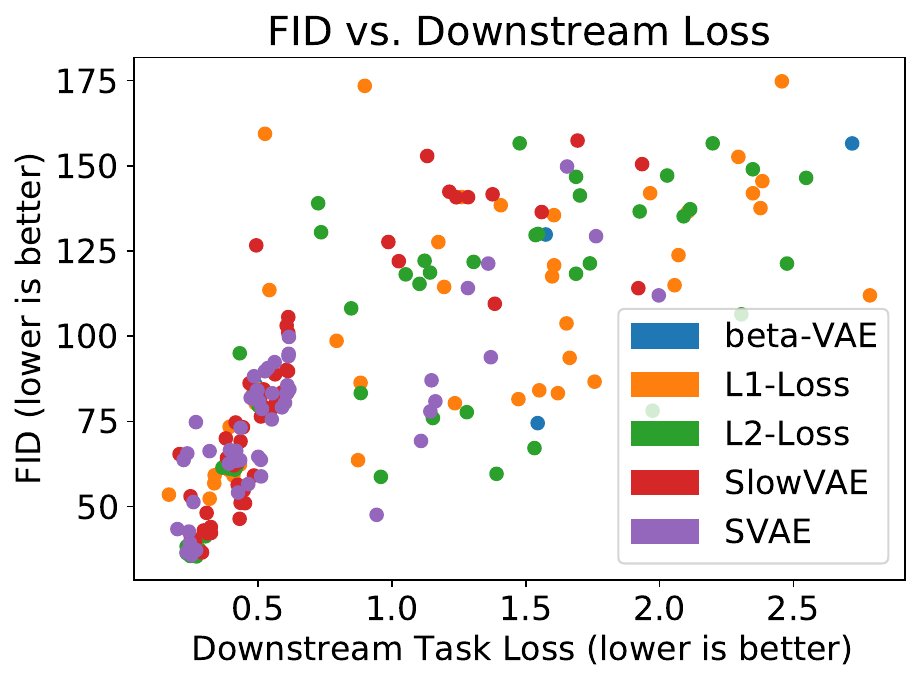}
\end{center}
\caption{Pong, $\frac{1}{4}$ of data}
\label{fig:pong_fid_part}
\end{subfigure}
\begin{subfigure}{0.32\linewidth}
\begin{center}
\includegraphics[width=\textwidth]{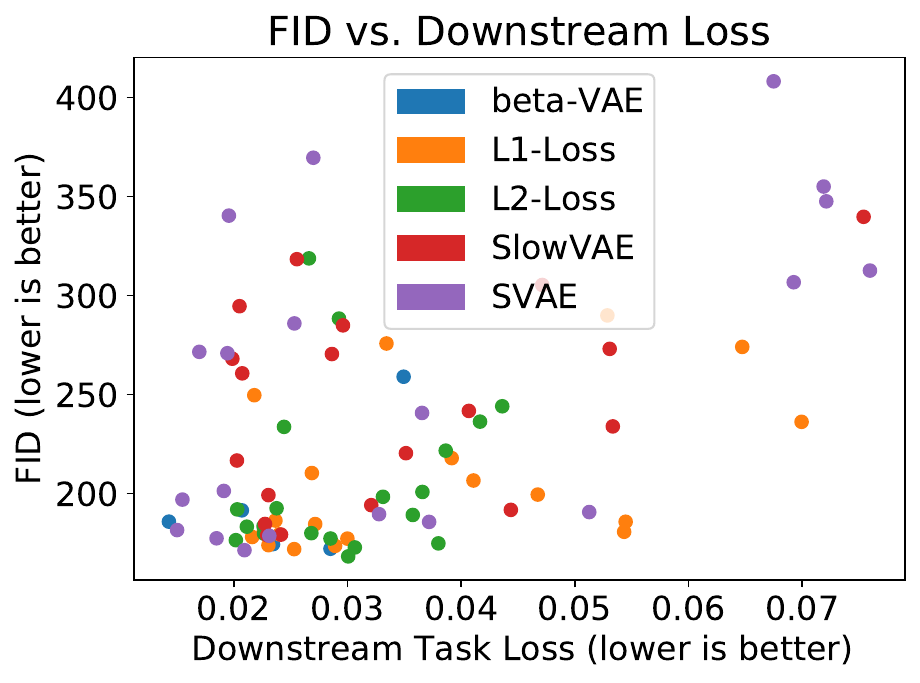}
\end{center}
\caption{DeepMind Lab, $\frac{1}{4}$ of data}
\label{fig:dml_fid_prt}
\end{subfigure}
\caption{Scatter plot visualization of downstream task performance vs. FID score of the autoencoding model colored by method.}
\label{fig:fid}
\end{figure*}
\section{Predicting Downstream Task Performance}
As we have seen the qualitative latent space properties discussed above do not give the necessary predictive power to select an encoder model with good downstream task performance.
To predict general downstream task performance, we want to study each of the three components of the general formulation in Eq. \eqref{eq:slow_highlevel}: reconstruction performance, disentanglement and slowness. 

The extent to which \textbf{disentanglement} can predict downstream task performances has been investigated by Locatello et al. in \cite{locatello2019challenging}.
In their work the authors questioned the benefits that disentangled representations have for learning downstream tasks and criticised that currently all disentanglement metrics require ground truth labels.
Since we assume sparse labeled data and want to predict performance for any downstream tasks, we are not able to use existing disentanglement metrics.

The second option is to measure if a latent space exhibits the \textbf{slowness} properties and correlate this measure with downstream task performance.
We devised a metric that measures the length of a trajectory in latent space defined by $N$ encoded latent representations $z_1, \ldots, z_N$ and compared it to the euclidean distance between $z_1$ and $z_N$.
According to the slowness principle and the qualitative analysis in Fig. \ref{fig:2d_latents} we hypothesize that higher slowness regularization with a qualitatively smoother latent space has less jumps and therefore the trajectory in latent space is shorter and more similar to the euclidean distance.
Based on a preliminary study we could observe that higher slowness regularization indeed leads to less fragmented and shorter latent trajectories.
However, the metric does not correlate with downstream task performance.

Third, we can investigate if the \textbf{generative capabilities} of a model allow us to predict downstream task performance.
The core idea is that a model capable of generating "realistic" images from representations sampled from the latent distribution encodes more useful information.
We use the FID which is usually used to evaluate the generative capabilities of Generative Adverserial Networks (GANs).
After training the autoencoder we draw random samples from the latent distributions and decode them into images.
Then we compute the FID between the training data distribution and the generated data distribution.
The results of this analysis on the three experiments in this paper are shown in Fig. \ref{fig:fid}.
Furthermore, Figure \ref{fig:latentanalysis} is showing the FID for all $\lambda$, $\beta$ combinations for the S-VAE.
For both cases, abundant and sparse labeled training data during the downstream task, we can see a correlation between low FID and low downstream task loss.
In the ball experiment the best performing $L2$ loss model can clearly be identified by its low FID.
In the other experiments it is not as obvious to find a clearly superior method, but in all experiments the FID can be used to identify a small set of good performing VAE models.
In conclusion, the FID score is a good measure to predict downstream task performance of an autoencoding model and therefore greatly improves hyperparamter tuning speed saving valuable time.

\section{Slow Autoencoding Methods from an Information Theoretic Point of View}
Lastly we want to discuss the connection between the FID score as a predictive metric for downstream task performance of a model and slowness regularization.
Let us look at the $\beta$-VAE from the Information theoretic point of view.
We can write the $\beta$-VAE objective as $\max[I(z;y) - \beta I(o;z)]$ \cite{burgess2018understanding,alemi2016deep} where $I(z;y)$ is the mutual information of the intermediate representation $z$ and the reconstruction task.
The subtracted term $\beta I(o;z)$, usually referred to as the latent bottleneck, effectively encourages the model to discard less relevant information present in the input $o$ by limiting the capacity of the latent information channels.
In our case such information would be properties of the input signal that are not relevant for reconstruction.
As observed in the previous analysis, increasing the latent bottleneck pressure by varying the parameter $\beta$ reduces the number of latent dimensions used.
As we have discussed in \ref{sec:hyperparams}, the influence of the slowness regularization on the latent bottleneck pressure remains somewhat unclear.
We theorize that, similar to how the $\beta$-VAE discards less relevant information, slowness regularization adds a way for the model to include or discard information when learning temporal tasks.
This draws a parallel to the core idea of the slowness principle: Discourage the encoding of irrelevant quickly changing components of the input signal and encourage encoding of slowly changing features.

\section{Conclusion}
In this paper, we propose a general framework for applying the slowness principle to the state-of-the-art $\beta$-VAE.
Within this framework we compare existing methods of enforcing slowness such as $L1$ and $L2$ loss and the SlowVAE.
Furthermore, we introduce a new slowness regularization term based on a Brownian motion.
We find that slow methods outperform the baseline $\beta$-VAE with respect to downstream task data efficiency and performance for experiments in simple dynamic environments.
The results indicate that $L1, L2$ slowness loss terms perform as good as the more complex slowness loss terms taking uncertainty into account.
A qualitative analysis of the latent space structure illustrates the effects of the slowness loss on the latent space structure.
Lastly we find that the Fr\'echet Inception Distance is a good measure to predict downstream task performance by quantifying how much of the encoded information in the latent space is useful for reconstructing images.

An important open question is, how the properties of the latent space influences the optimization of downstream tasks.
The results indicate that a smooth latent space is not always the best for learning downstream tasks and that there is some trade-off between the slowness regularization and latent bottleneck pressure.

\bibliographystyle{named}
\bibliography{ijcai21}

\appendix

\section{Experiment Details}\label{app:experiments}
All experiments have been implemented in the same VAE framework.
We used the Adam optimizer as introduced in \cite{kingma2014adam}.
The reconstruction loss is computed using the binary cross entropy between predicted and true images.
\subsection{Ball Experiment}
The goal in this task is to predict the velocity of the ball from two consecutive frames.
We generated sequences of 20 frames of a ball bouncing in a $100 \times 100$ uni-color 2D environment.
For each sequence the ball is placed in a random position in the environment and initialized with a random direction and random (within some limits) velocity vector.
Upon reaching the border of the environment, the velocity vector of the ball is flipped to mimic the principle "incident angle equals emergence angle" while keeping the balls velocity constant.
The environment performs 20 update steps by applying the velocity vector to the ball and stores the frames and the ball's x/y-velocity.
Overall 10000 labeled sequences consisting of 20 datapoints each were generated.

During the \textbf{unsupervised representation learning step}, we use the full dataset without labels to train multiple models of all methods by varying the hyperparameters $\beta$ and $\lambda$.
The training consisted of 150 epochs with a batch size of 16 and the learning rate $1e-03$.
The encoder consists of 3 fully connected layers of size 100x100, 300 and 2 (latent dimensions). Mean and log-variance of the latent distributions is computed using two more layers with size 2.
The decoder is a mirrored version of the encoder without the two mean and log-variance layers.

In the \textbf{supervised downstream task} we use subsets of ($1, 1/2, 1/4, \ldots, 1/128$) of the full labeled dataset with their labels to train the downstream task of predicting the ball velocity from two consecutive frames.
The downstream task is trained by freezing the encoder networks trained in the previous step and feeding the latent representations into a fully connected neural network with $2$ layers of $50$ hidden nodes.
The downstream loss is computed as the MSE between true and predicted velocity vector.
Overall 140 models were trained.

\subsection{Pong experiment}
The goal was to predict the actions that both agents are taking from two consecutive frames.
A dataset consisting of 5000 sequences with 20 data points each was collected by observing two agents play the game of Pong against each other.
Sequences that contained reset events of the environment, for example when a game was won/lost were discarded during dataset generation to obtain uninterrupted sequences..
The training of this experiment was conducted similar to the ball experiment, obtaining two consecutive latent representations, concatenating them and obtaining the predicted action probabilities.
Subsets of ($1, 1/2, 1/4, \ldots, 1/64$) of the full labeled dataset with their labels were used to train the downstream task.
The same 2 layer, 50 nodes downstream neural network structure was used.
The loss is computed as the Binary Cross Entropy loss between the predicted and one-hot encoded action probabilities.
The encoder network consists of 4 convolutional layers with batch normalization followed by 2 fully connected layers with 1024 and 512 hidden nodes.
The decoder network mirrors the encoder network.
In this experiment a batch size of 32 was used and the training was performed for 250 epochs.

\subsection{Deepmind Lab Dataset}
In this experiment the goal is to learn the $6$-DOF visual odometry of an agent exploring a DeepMind Lab environment~\cite{beattie2016deepmind}.
A dataset of 19000 sequences with 20 data points each was generated using a random walker exploring the environment.
Subsets of ($1, 1/2, 1/4, \ldots, 1/32$) of the full labeled dataset with their labels were used to train the downstream task.
The same 2 layer, 50 nodes downstream neural network structure was used.
The loss is computed as the MSE loss between the predicted and true $6D$ motion vector.
The encoder network consists of 6 convolutional layers with 1 intermediate residual block for each layer followed by a full connected layer with 8192 hidden nodes.
The decoder network mirrors the encoder network.

\section{Performance Result Details}\label{app:performance}
The mean downstream performance for all trained models on the full and $\frac{1}{4}$ downstream dataset is summarized in Appendix \ref{app:performance}.
In the tables \ref{tab:blob_data} to \ref{tab:dml_data} we can see that in most cases with abundant data the $L2$ slowness loss term leads to the best or comparable performance to the S-VAE and SlowVAE.
In the pong experiment, the S-VAE and SlowVAE performed both better compared to the $L1$ and $L2$ methods when labeled data was sparse.
Overall there seems to be little difference between the simpler $L1$ and $L2$ methods and also between S-VAE and SlowVAE which take the covariances of the latent distributions into account.
\begin{table}
    \centering
    \caption{Mean and STD downstream task loss for all slow methods in the ball experiment.The best performing methods are highlighted in blue.}\label{tab:blob_data}
    \begin{tabular}{r|ll|ll}
      \toprule 
      & Abundant data & & Sparse data & \\
      \bfseries Method & \bfseries Mean & \bfseries STD & \bfseries Mean & \bfseries STD\\
      \midrule 
      S-VAE & 1.259 & 0.447 & 1.464 & 0.408\\
      SlowVAE & \textcolor{blue}{1.095} & \textcolor{blue}{0.219} & 1.339 & 0.219\\
      $L1$ & 1.310 & 0.441 & 1.525 & 0.422\\
      $L2$ & 1.153 & 0.430 & \textcolor{blue}{1.292} & \textcolor{blue}{0.438}\\
      \bottomrule 
    \end{tabular}
\end{table}
\begin{table}
    \centering
    \caption{Mean and STD downstream task loss for all slow methods in the pong experiment.The best performing methods are highlighted in blue.}\label{tab:pong_data}
    \begin{tabular}{r|ll|ll}
      \toprule 
      & Abundant data & & Sparse data & \\
      \bfseries Method & \bfseries Mean & \bfseries STD & \bfseries Mean & \bfseries STD\\
      \midrule 
      S-VAE & 0.376 & 0.139 & \textcolor{blue}{0.657} & \textcolor{blue}{0.440}\\
      SlowVAE & 0.373 & 0.126 & \textcolor{blue}{0.650} & \textcolor{blue}{0.427}\\
      $L1$ & \textcolor{blue}{0.332} & \textcolor{blue}{0.078} & 1.034 & 0.752\\
      $L2$ & \textcolor{blue}{0.326} & \textcolor{blue}{0.079} & 1.040 & 0.707\\
      \bottomrule 
    \end{tabular}
\end{table}
\begin{table}
    \centering
    \caption{Mean and STD downstream task loss for all slow methods in the DeepMind Lab experiment. The best performing methods are highlighted in blue.}\label{tab:dml_data}
    \begin{tabular}{r|ll|ll}
      \toprule 
      & Abundant data & & Sparse data & \\
      \bfseries Method & \bfseries Mean & \bfseries STD & \bfseries Mean & \bfseries STD\\
      \midrule 
      S-VAE & 0.031 & 0.021 & 0.035 & 0.022\\
      SlowVAE & 0.028 & 0.014 & 0.032 & 0.014\\
      L1 & 0.026 & 0.020 & 0.033 & 0.015\\
      $L2$ & \textcolor{blue}{0.021} & \textcolor{blue}{0.010} & \textcolor{blue}{0.029} & \textcolor{blue}{0.007}\\
      \bottomrule 
    \end{tabular}
\end{table}

\section{Analysis}\label{app:analysis}
\begin{figure*}
\begin{subfigure}{0.245\linewidth}
\begin{center}
\includegraphics[width=\textwidth,trim={0cm 0cm 0cm 0.8cm},clip]{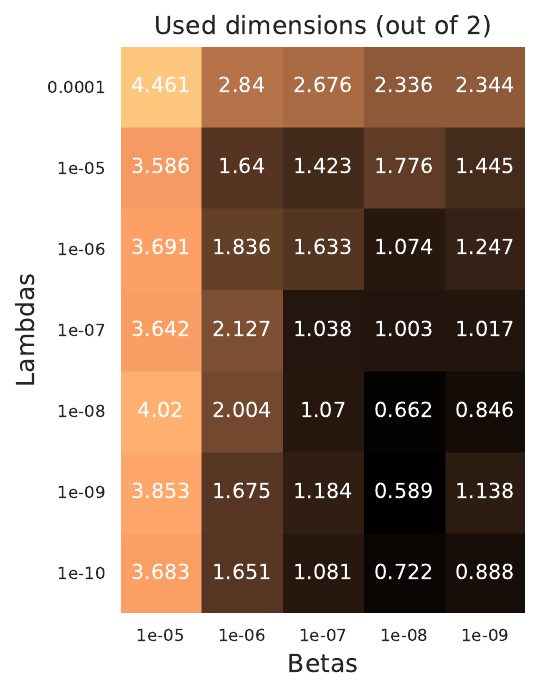}
\end{center}
\caption{Ball experiment, downstream task performance full labeled data, method SVA.}
\label{fig:ball_overview_full}
\end{subfigure}
\begin{subfigure}{0.245\linewidth}
\begin{center}
\includegraphics[width=\textwidth,trim={0.7cm 0cm 0cm 0.8cm},clip]{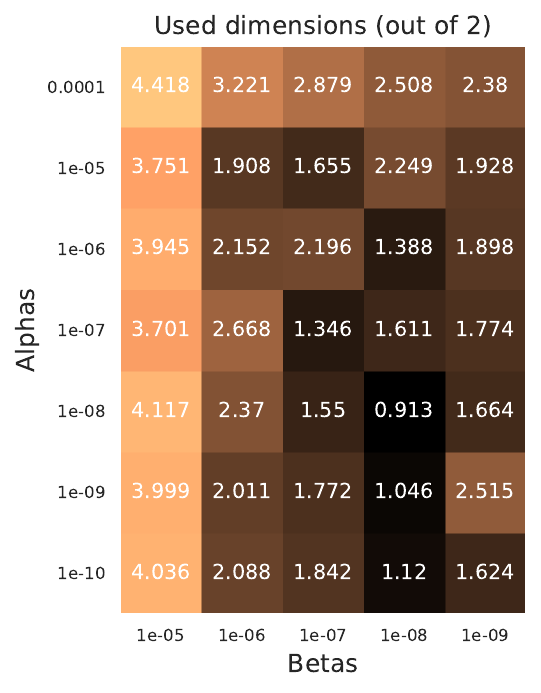}
\end{center}
\caption{Ball experiment, downstream task performance $\frac{1}{8}$ of the labeled data, method SVA.}
\label{fig:ball_overview_part}
\end{subfigure}
\begin{subfigure}{0.245\linewidth}
\begin{center}
\includegraphics[width=\textwidth,trim={3.5cm 0cm 2.5cm 1cm},clip]{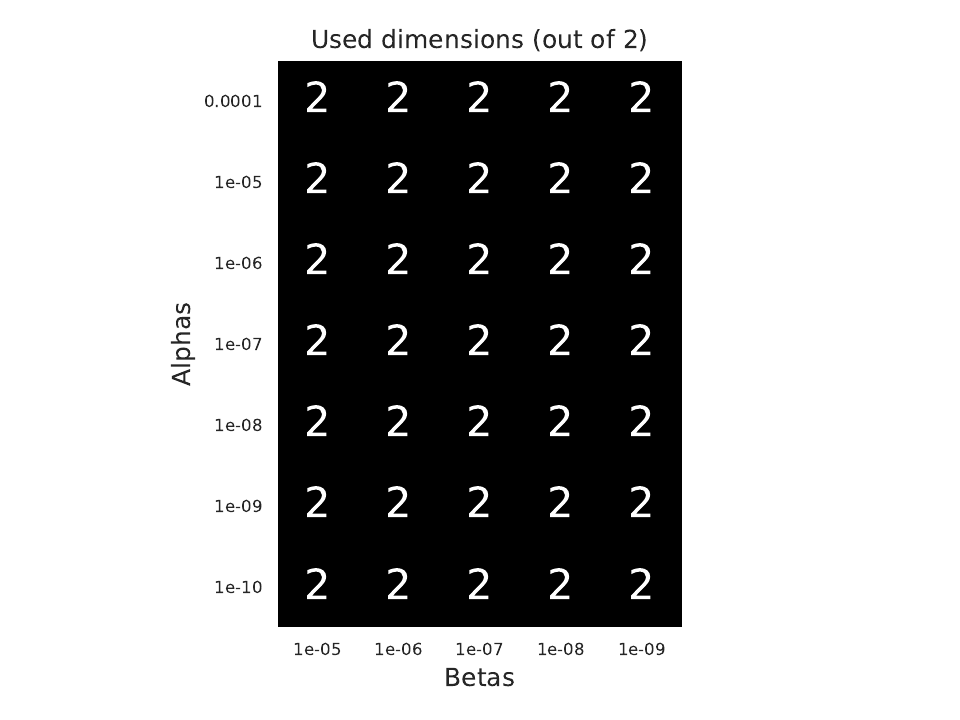}
\end{center}
\caption{Ball experiment, used dimension out of 2, method S-VAE}
\label{fig:ball_overview_used}
\end{subfigure}
\begin{subfigure}{0.245\linewidth}
\begin{center}
\includegraphics[width=\textwidth,trim={0.76cm 0cm 0cm 0.8cm},clip]{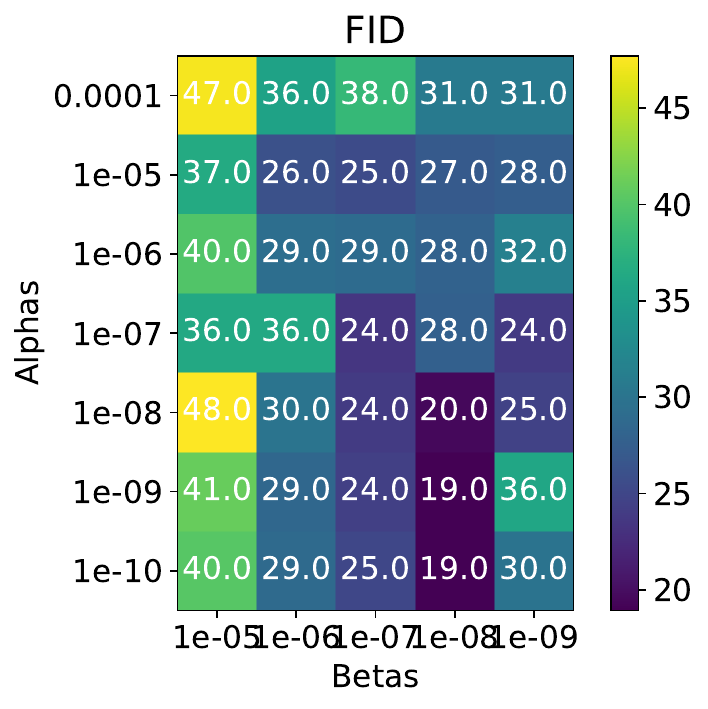}
\end{center}
\caption{Ball experiment FID scores, method SVAE}
\label{fig:ball_overview_fid}
\end{subfigure}
\\
\begin{subfigure}{0.245\linewidth}
\begin{center}
\includegraphics[width=\textwidth,trim={0cm 0cm 0cm 0.8cm},clip]{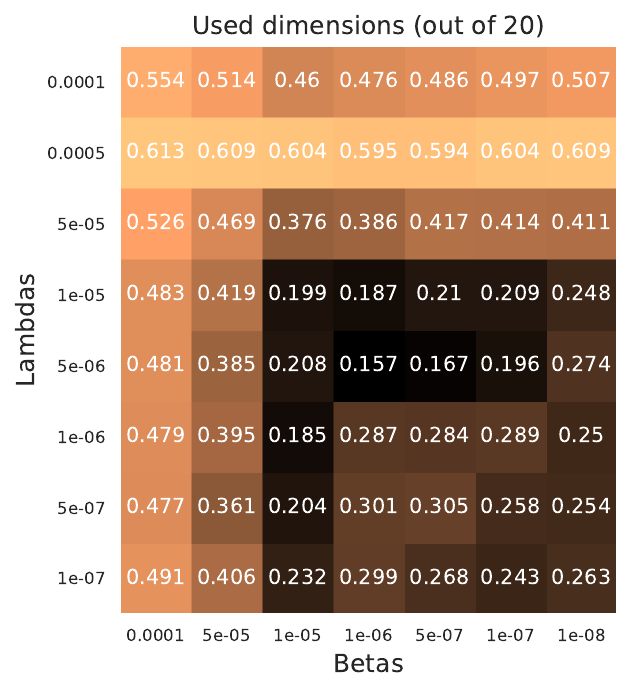}
\end{center}
\caption{Pong experiment, downstream task performance full labeled data, method S-VAE.}
\label{fig:pong_overview_full}
\end{subfigure}
\begin{subfigure}{0.245\linewidth}
\begin{center}
\includegraphics[width=\textwidth,trim={1cm 0cm 0cm 0.8cm},clip]{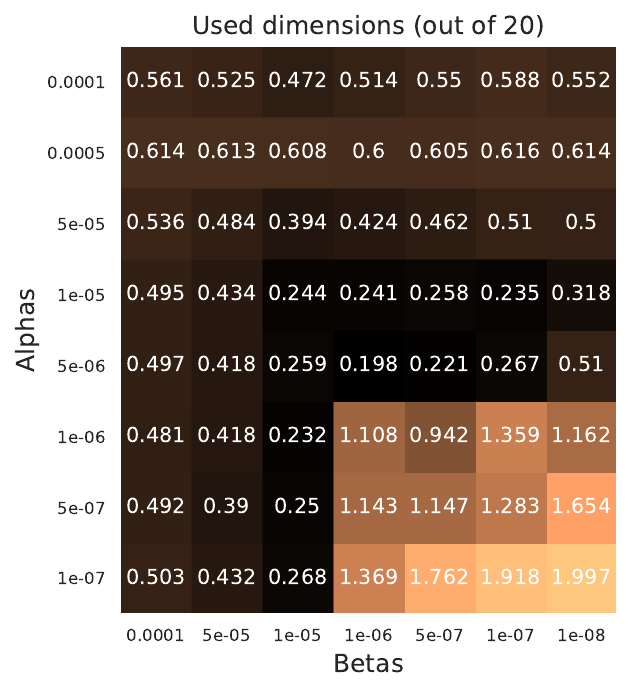}
\end{center}
\caption{Pong experiment, downstream task performance $\frac{1}{4}$ of the labeled data, method S-VAE.}
\label{fig:pong_overview_part}
\end{subfigure}
\begin{subfigure}{0.245\linewidth}
\begin{center}
\includegraphics[width=\textwidth,trim={2.75cm 0cm 2.5cm 1cm},clip]{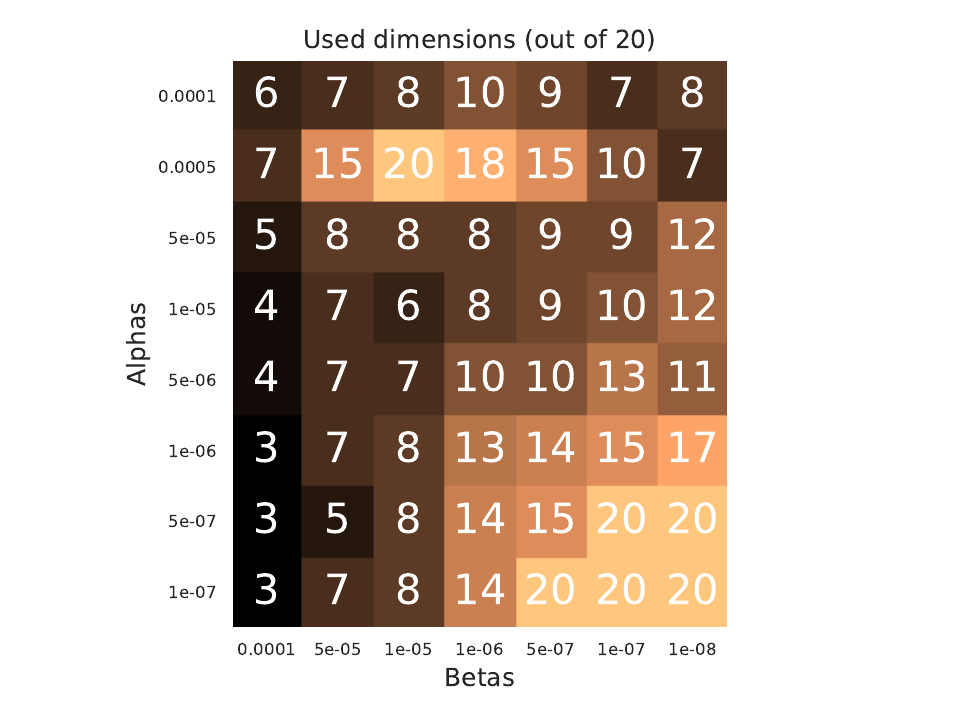}
\end{center}
\caption{Pong experiment, used dimensions out of 20, method SVAE}
\label{fig:pong_overview_used}
\end{subfigure}
\begin{subfigure}{0.245\linewidth}
\begin{center}
\includegraphics[width=\textwidth,trim={0.6cm 0cm 0cm 0.75cm},clip]{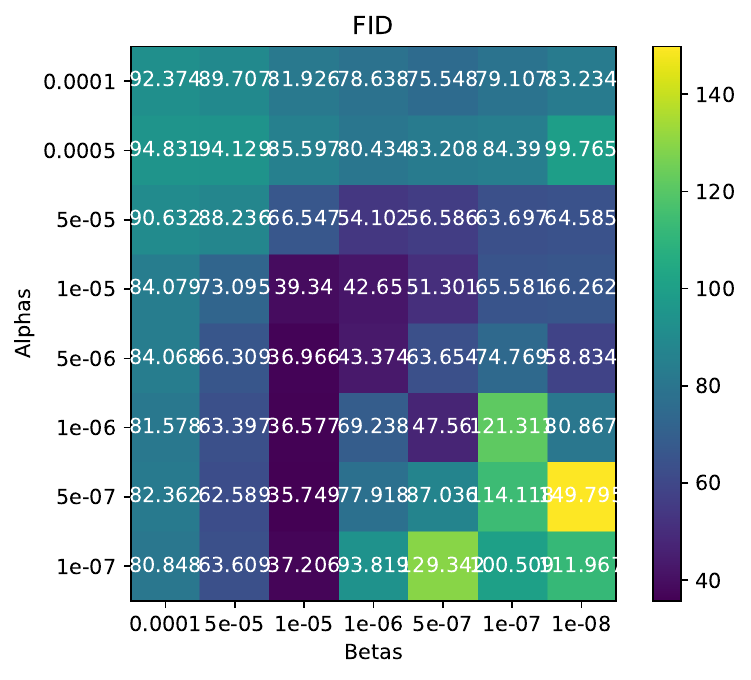}
\end{center}
\caption{Pong experiment FID scores, method, SVAE}
\label{fig:pong_overview_fid}
\end{subfigure}
\\
\begin{subfigure}{0.245\linewidth}
\begin{center}
\includegraphics[width=\textwidth,trim={0cm 0cm 0cm 0.8cm},clip]{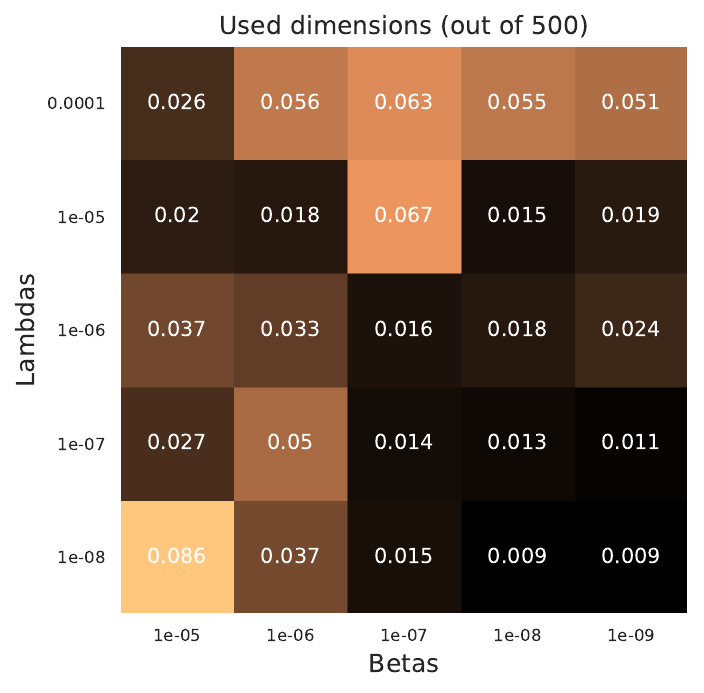}
\end{center}
\caption{DeepMind Lab experiment, downstream task performance full labeled data, method S-VAE.}
\label{fig:dml_overview_full}
\end{subfigure}
\begin{subfigure}{0.245\linewidth}
\begin{center}
\includegraphics[width=\textwidth,trim={0.6cm 0cm 0cm 0.8cm},clip]{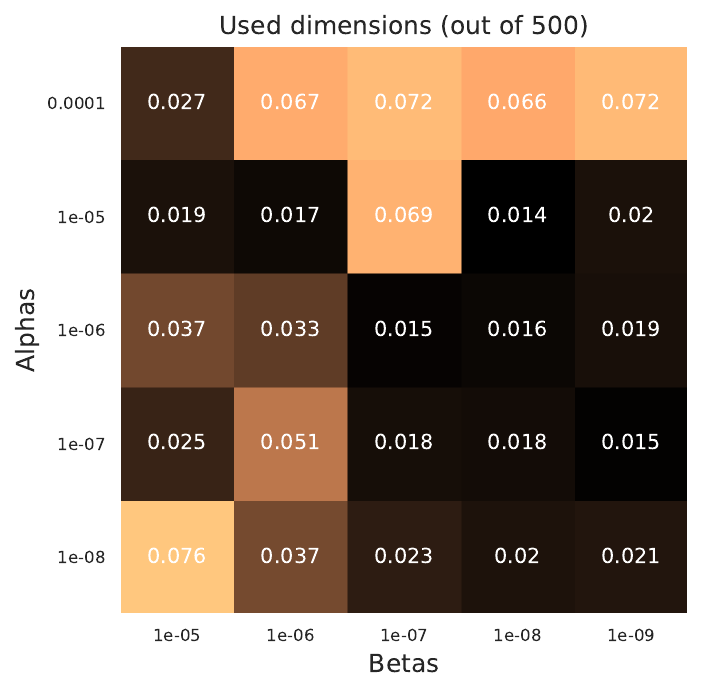}
\end{center}
\caption{DeepMind Lab experiment, downstream task performance $\frac{1}{4}$ of the labeled data, method S-VAE.}
\label{fig:dml_overview_part}
\end{subfigure}
\begin{subfigure}{0.245\linewidth}
\begin{center}
\includegraphics[width=\textwidth,trim={2cm 0cm 2.5cm 1cm},clip]{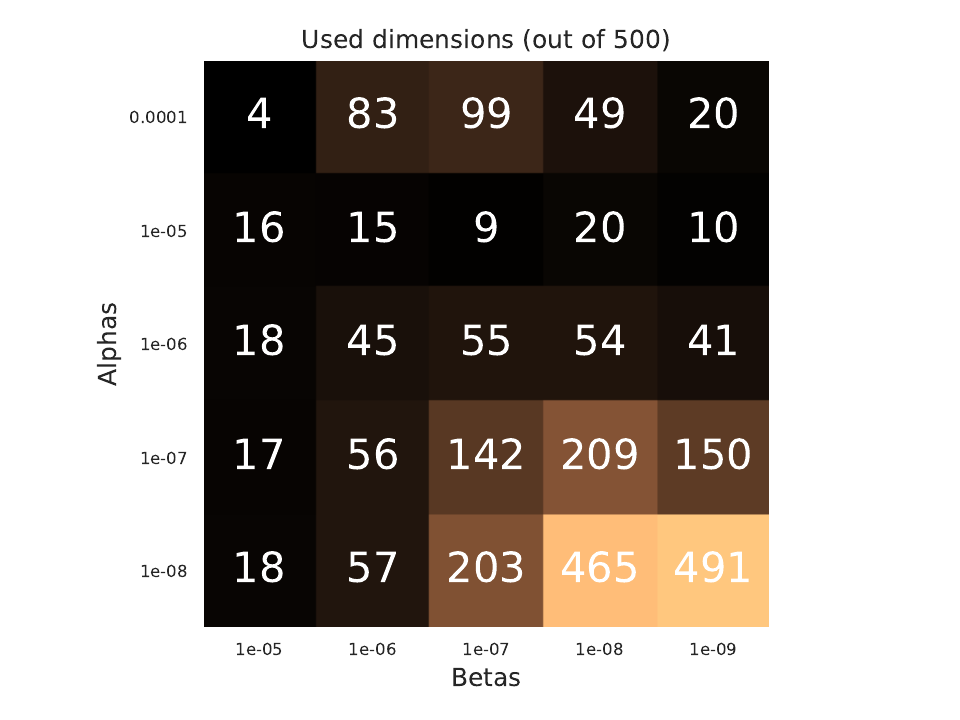}
\end{center}
\caption{DeepMind Lab experiment, used dimensions out of 500, method S-VAE}
\label{fig:dml_overview_used}
\end{subfigure}
\begin{subfigure}{0.245\linewidth}
\begin{center}
\includegraphics[width=\textwidth,trim={1cm 0cm 0cm 0.8cm},clip]{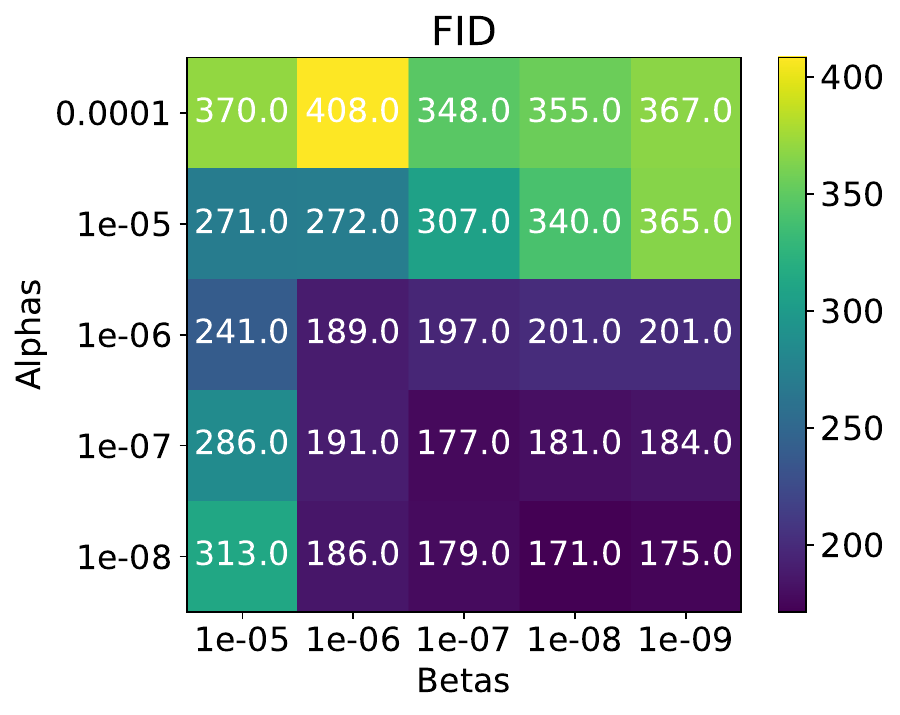}
\end{center}
\caption{DeepMind Lab experiment FID scores, method, S-VAE}
\label{fig:dml_overview_fid}
\end{subfigure}
\caption{Latent space analysis for all three experiments. From top to bottom: Ball experiment, Pong Experiment DeepMind Lab experiment. From left to right: downstream task performance with full and partial labeled training datasets, used dimensions, FID and interpolation scores.}
\label{fig:latentanalysis}
\end{figure*}

In Fig. \ref{fig:latentanalysis} we summarize the results of all three experiments (from top to bottom: Ball experiment, Pong experiment, DeepMind Lab experiment) for the S-VAE.
From left to right we plot different metrics for all models trained (different combinations of hyperparameters $\lambda$ and $\beta$:
First, the downstream task performance for the full and partial labeled datasets.
Then the number of latent channels used by each model and the FID.

\end{document}